\documentclass{article}
\usepackage[table]{xcolor}
\usepackage{graphicx}
\graphicspath{ {./resources/} }
\usepackage{pdfpages}
\usepackage{caption}
\usepackage{subcaption}
\usepackage{float}
\usepackage{amsmath}
\usepackage{hyperref}
\hypersetup{
  colorlinks,
  citecolor=blue,
  linkcolor=blue,
  urlcolor=blue
}
\usepackage{amsfonts}
\usepackage{authblk}
\usepackage[inline]{enumitem}
\usepackage{lscape}
\usepackage{adjustbox}

\title{Model-assisted Reinforcement Learning of a Quadrotor}
\date{}

\author[1]{Arshad Javeed}

\affil[1]
{
Dept. of Automatic Control,

Lund University
}

\begin{document}
\maketitle

\begin{abstract}
	In recent times, reinforcement learning has produced baffling results when it comes to performing control tasks with highly non-linear systems. The impressive results always outweigh the potential vulnerabilities or uncertainties associated with the agents when deployed in the real-world. While the performance is remarkable compared to the classical control algorithms, the reinforcement learning-based methods suffer from two flaws, robustness and interpretability, which are vital for contemporary real-world applications. The paper attempts to alleviate such problems with reinforcement learning and proposes the concept of ``model-assisted'' reinforcement learning to induce a notion of conservativeness in the agents. The control task considered for the experiment involves navigating a CrazyFlie quadrotor. The paper also describes a way of reformulating the task to have the flexibility of tuning the level of conservativeness via multi-objective reinforcement learning. The results include a comparison of the vanilla reinforcement learning approaches and the proposed approach. The metrics are evaluated by systematically injecting disturbances to classify the inherent robustness and conservativeness of the agents. More concrete arguments are made by computing and comparing the backward reachability tubes of the RL policies by solving the Hamilton-Jacobi-Bellman partial differential equation (HJ PDE).
\end{abstract}

\section{Introduction}

The classical control approach relies on the ability to accurately model the system and leverage a proven and robust control algorithm. The challenge here is coming up with a decently accurate mathematical description of the system, and the problem can be exacerbated in the case of a non-linear or a high-dimensional system. In this regard, the reinforcement learning approach, with the machine-learning methodology of learning from the data has yielded unmatched results in diverse application domains while requiring a minimal description of the system \cite{Kober2014, razzaghi2022survey, Bai_2023, Arulkumaran_2017}.

Reinforcement learning is closely related to optimal control \cite{Sutton1998}. In the context of control, the agent tries to learn an optimal control policy by interacting with the environment by receiving rewards for its actions. The fundamental idea is to maximize the expected reward received from the environment. The learning can be model-based or model-free. The contemporary model-based and model-free algorithms are enriched by deep learning and neural networks acting as non-linear function approximators. The ``model'' here refers to the model of the environment. In an MDP setting, the model-free approach does not leverage the dynamics of the world in any way, instead tries to approximate the reward function and the cost function by repeated interaction. There are several algorithms in the model-free category. \cite{mnih2016asynchronous, schulman2017proximal} directly try to optimize the policy function parameterized by a deep neural network. \cite{sac, ddpg, td3} are actor-critic variants that employ neural networks for the actor and the critic, where the critic approximates the true cost function using the recursive Bellman equation. Model-based reinforcement learning, on the other hand, uses the dynamics (predefined or learns simultaneously) to define the state transition probabilities to seek better actions and reduce repeated trial and error in the environment, and the state transition probabilities can be utilized by the algorithm in several ways. MBVE \cite{mbve} uses model-based learning for better estimation of the value function. MBMF \cite{mbmf} first learns the model or the environment dynamics using a shallow neural network and then uses a Model Predictive Control (MPC). Expert Iteration \cite{expert} uses an expert improvement step to define the imitation learning targets and then resolves to imitation learning for faster convergence. 

With reinforcement algorithms employed in diverse applications, robustness and interpretability are imperative. When employed in real-world applications, any reckless behavior can have ramifications for the agent itself or the surroundings. There have been several attempts to incorporate safety within reinforcement learning. \cite{srinivasan2020learning} employ a safety critic that evaluates the safety of the given state-action pair looking at the unsafe experiences of the agent and then transforms the task to safety-constrained MDP. Since the unsafe conditions are determined by the data, this may induce a bias. The constraints themselves are less interpretable due to the fact that a black box model or neural network tries to learn from them. SAVED \cite{saved} is a deep MPC that learns the model by training neural networks observing suboptimal demonstrations. The learned model is then used to impose constraints on MPC optimization. \cite{brunke2021safe} is a comprehensive review of safety-constrained reinforcement learning. MPC approach is one of the only non-linear control strategies that can guarantee the constraint requirements, the only downside is that it is computationally expensive.

Broaching the topic of robustness, ``robustness'' can be defined as the ability of the model to perform consistently (or even better than expected) in diverse situations. Robustness does not necessarily imply that the agent performs better in unknown environments. On the contrary, we want the agent to honor certain physical constraints whilst compromising the reward objective, but ultimately accomplishing the task. In the context of reinforcement learning, this could be quite tricky, as often conflicting objectives are represented by the reward signal. \cite{glossop2022characterising} have evaluated the robustness of reinforcement learning algorithms by systematic disturbance injection. A startling discovery made by them is that subjecting the agent to disturbances during the training phase does not necessarily improve the robustness, i.e. the inherent robustness of agents trained with and without disturbances is about the same. \cite{MAYNE2005219} have solved the classical model predictive control approach by incorporating bounded disturbances during online optimization. \cite{morimoto} propose an adversarial setting (robust reinforcement learning), where an adversarial agent modeling the environment tries to impede the performance, while the RL agent performing the task tries to overcome it. However, \cite{glossop2022characterising} report that the robust reinforcement learning methods do not considerably outperform the baseline model.

In contrast, the proposed work aims to use a partially specified model or learn a partial model and then use the model constraints as one of the reward objectives with the intention of reinforcement learning agent learning to act in a way that would honor the model constraints. Often in reinforcement learning, there are conflicting objectives and there is always an added advantage of taking the risk. The primary motivation of the design is the ability to individually tune the constraints during run-time by employing multi-objective reinforcement learning (depending on the user preferences). Similar to \cite{glossop2022characterising}, the evaluation is done by systematically injecting disturbances during the test phase and looking at the agent's capability to adhere to the defined constraints. The proposed work corroborates the implicit robustness of reinforcement learning agents, i.e. training with disturbance does not necessarily improve the reward. But at the same time, discusses interesting properties of the proposed methodology. All the experiments are performed on a highly non-linear real-world system, a CrazyFlie quadrotor. More details of the task are presented in the subsequent sections.

\section{Background}

Reinforcement learning is a type of unsupervised machine learning, where the agent learns an optimal policy for the given task by interacting with the environment. The ``feedback signal'' or the reward signal is the key to reinforcement learning. During the training process, the agent's goal is to simply learn a strategy that would maximize the cumulative reward.

At any given time $t$, the agent could be in an arbitrary state $s_t \in \mathcal{S}$ and can choose to perform an action $a_t \in \mathcal{A}$. As a consequence of the action, the agent receives a reward $R(s_t, a_r) = r_t \in \mathbb{R}$. So the cumulative reward for acting according to the policy $\pi$ starting from state $s_t$ is $\Sigma_{t'=t}^\infty \gamma^{t' - t} R(s_{t'}, \pi(s_{t'}))$, where $\gamma \in [0, 1)$ is called the discount factor. Thus, having defined a reward function $R$, the goal is to learn an optimal policy $\pi^*$ that maps the states to actions $\pi^* : \mathcal{S} \to \mathcal{A}$ that yields the best discounted reward.

If the above task is done iteratively without assuming any knowledge of the system or the dynamics of the environment, it is called ``model-free'' reinforcement learning. On the other hand, if the information about the dynamics or the state transitions is leveraged or learned during the process, it is ``model-based learning'' RL. The transition probabilities here are the conditional probabilities $P(s_{t+1} | s_t, a_t)$, which resemble the transition probabilities typically encountered in a Markov decision problem (MDP).

In the above formulation, the reward $r_t$ was assumed to be a scalar. Practical applications (as well as in the scope of the paper) deal with the tradeoff between multiple constraints, and there are often conflicting objectives or rewards. Traditionally, a utility function is used to scalarize the reward. A utility function scalarizes the reward as $u : \mathbb{R}^d \to \mathbb{R}$ and the discounted reward becomes $\Sigma_{t'=t}^\infty \gamma^{t' - t} u(r_{t'})$, and the value function then corresponds to Expected Scalarized Return (ESR). A common utility function is simply the weighted average, $u(r_t) = \Sigma_i w_i r_{ti}$. However, this would muddle the reward objectives and make them less interpretable. An alternative approach alluding to multi-objective reinforcement learning is the Scalarized Expected Return (SER) approach, where the utility function is applied after the inner expectation, i.e. $ u\left( \Sigma_{t'=t}^\infty \gamma^{t' - t} r_{t'} \right)$. In multi-objective reinforcement learning, the goal is now to learn an array of policies each of which prioritizes the reward objectives differently. \cite{Hayes_2022} provide a detailed overview of the strategies involved in multi-objective learning.

The usage of deep neural networks as function approximators to approximate the policies $\pi : \mathbb{R}^{n_S} \to \mathbb{R}^{n_A}$ and value functions $Q : \mathbb{R}^{n_S} \times \mathbb{R}^{n_A} \to \mathbb{R}^{n_R}$ makes it feasible to model non-linear dynamical systems and learn effective control strategies \cite{Arulkumaran_2017, Ladosz_2022, 9023331}.

\section{Formulating the RL Task}

\begin{figure}[h]
	\centering
	\caption{Navigation Task}
	\includegraphics[scale=0.4]{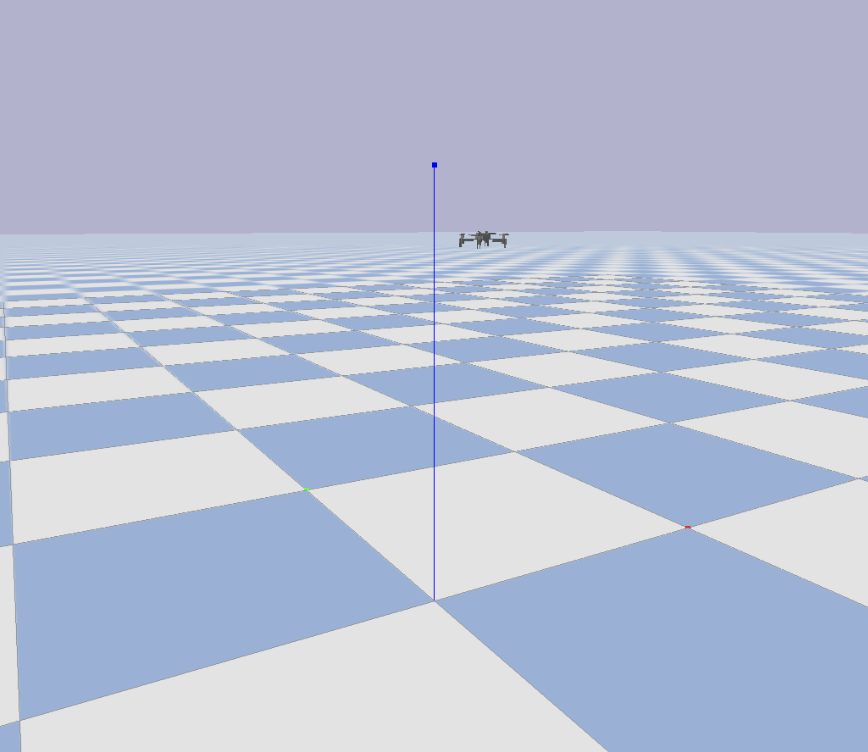}
	\label{nav_task:fig}
\end{figure}

The non-linear dynamical system considered here is a CrazyFlie quadrotor. The task for the reinforcement learning agent is to learn a policy to navigate the quadrotor to the specified destination starting from an initial state $s_0$ (figure \ref{nav_task:fig}). However, the environment poses a challenge to the agent, there are always disturbances (forces emulating wind) acting along the $X, Y, Z$ directions randomly. The observation space or the state space consists of the system kinematics (equation \ref{obs_space:eq}), involving the 3D position information, orientation (roll, pitch, and yaw), and linear and angular velocities along the $X, Y, Z$ axes. The state space $S$ will later be extended with additional states later on.

\begin{equation}
	\label{obs_space:eq}
	\mathcal{S} = [x, y, z, r, p, y, v_x, v_y, v_z, w_x, w_y, w_z]
\end{equation}

Now, there are several ways of formulating the action space for the navigation task. We can have the reinforcement learning agent have complete granular control of the quadrotor by letting it control the low-level rotor RPMs or use a hierarchical control approach. The hierarchical approach involves the reinforcement learning agent defining high-level objectives for the quadrotor and the quadrotor employing a low-level controller such as a PID controller to execute the actions. Employing a trusted low-level controller has a twofold advantage. Firstly, it means that the RL agent does not have to learn everything from scratch, for instance, simple maneuvers like moving up, down, diagonally, etc. can be performed by the low-level controller confidently, while the high-level controller is only concerned about taking the appropriate course of actions. This significantly alleviates the algorithmic convergence. Secondly, the overall model is more interpretable, due to the fact that the high-level actions can be interpreted easily. \cite{panerati2021learning} is a simulation bed supporting both approaches. However, the latter approach is run in an open-loop fashion, i.e. given a target position, the PID controller is used to compute the corresponding rotor RPMs and the implementation involves applying the RPMs for a fixed number of physics steps, but this is flawed in the way that it does not necessarily guarantee that the quadrotor executes the actions successfully, i.e. it may fall short or be far off the target. In contrast, the proposed approach executes a closed-loop PID control action to ensure that the quadrotor successfully executes the high-level action to reasonable accuracy. This makes the overall action execution even more interpretable and significantly impacts the algorithmic convergence. Additionally, the action also includes a target velocity for the agent, to help compensate for the effect of wind by attempting to move faster.

Thus, given a state $s_t = [x, y, z, r, p, y, v_x, v_y, v_z,  \allowbreak w_x, w_y, w_z]$ the action space for the reinforcement learning agent consists of predicting the relative coordinates and the target velocity, i.e. $\mathcal{A} = [\Delta x, \Delta y, \Delta z, v_{target}]$. To avoid taking large strides leading to potentially dangerous situations, the relative coordinates have an upper limit (equation \ref{a_limits:eq}). This essentially means that given a state $s_t$, the agent can move anywhere within a cube of 0.05 m, traveling a distance of $\sqrt{|\Delta x|^2 + |\Delta y|^2 + |\Delta z|^2}$. Additionally, the target velocity $v_{target}$ is a scalar quantity with which the agent desires to move to the next location. The individual velocity components along $X, Y, Z$ directions are computed using equation \ref{v_target:eq}, the absolute value ensures that the target velocities stay positive and the velocities are clipped by a safe upper limit $v_{max}$.

\begin{equation}
	\label{a_limits:eq}
	|\Delta x| \leq 0.05, |\Delta y| \leq 0.05, |\Delta z| \leq 0.05
\end{equation}

\begin{equation}
	\label{v_target:eq}
	v = \begin{bmatrix}
		v_x \\
		v_y \\
		v_z \\
		\end{bmatrix}
		= clip \left(
		v_{target} * \begin{bmatrix}
		|\Delta x| \\
		|\Delta y| \\
		|\Delta z| \\
		\end{bmatrix}, v_{max}
		 \right)
\end{equation}

Thus, given the current state vector $s_t = [x, y, z, r, p, y, v_x, v_y, v_z, w_x, w_y, w_z]$ and an action $a_t = [\Delta x, \Delta y, \Delta z]$, the ideal next state to be in would be $s_{t+1} = [x + \Delta x, y + \Delta y, z + \Delta z, r', p', y', v_x', v_y', v_z', w_x', w_y', w_z']$. Defining the action space this way serendipitously enables us to define a partial state transition model (equation \ref{state_trans:eq}), where the covariance matrix of the normal distributions is assumed to be the deviations from the ideal state due to the wind disturbances acting on the system. In an ideal scenario, the PID control would ensure that the quadrotor reaches the expected mean of the normal distribution.

\begin{equation}
	\label{state_trans:eq}
	\begin{bmatrix}
		X_{t+1} \\
		Y_{t+1} \\
		Z_{t+1} \\
		\end{bmatrix} \sim
		\mathcal{N}
		\left( 
		\begin{bmatrix}
		x_t + \Delta x_t \\
		y_t + \Delta y_t \\
		z_t + \Delta z_t \\
		\end{bmatrix},
		\Sigma_{t+1}
		\right)
\end{equation}

Equation \ref{state_trans:eq} now serves as a partial state transition probability model $P(s_{t+1} | s_t, a_t)$, which can be thought of as additional set constraints (equations \ref{c1:eq} - \ref{c3:eq}) for the RL agent, i.e. the ability to successfully execute the actions by reaching the ideal expected next state. We can now augment our original system state-space with 3 new quantities $x_e, y_e, z_e$ serving as a notion of ``action errors'' along $X, Y, Z$ directions (equation \ref{new_state:eq}).

\begin{equation}
	\label{new_state:eq}
	\mathcal{S}_e = [\mathcal{S}; \ x_e, \ y_e, \ z_e]
\end{equation}

\begin{equation}
	\label{c1:eq}
	x_{t+1} = x_t + \Delta x
\end{equation}

\begin{equation}
	\label{c2:eq}
	y_{t+1} = y_t + \Delta y
\end{equation}

\begin{equation}
	\label{c3:eq}
	z_{t+1} = z_t + \Delta z
\end{equation}

As the task involves reaching a predefined destination $s_{d} = [x_{d}, y_{d}, z_{d}]$, one of the obvious reward objectives (given a state action pair $(s_t, \pi(s_t))$) could be the negative of squared distance from the destination. Equation \ref{rew1:eq} concerns the navigation part of the task. Thus the objective is to maximize the reward, with zero being the highest reward.

\begin{equation}
	\label{rew1:eq}
	R_{nav}(s_t, \pi(s_t)) = - \left( (x_{t+1} - x_d)^2 + (y_{t+1} - y_d)^2 + (z_{t+1} - z_d)^2 \right)
\end{equation}

\begin{equation}
	\label{rew2:eq}
	R_{e}(s_t, \pi(s_t)) = - \left( x_e^2 + y_e^2 + z_e^2 \right)
\end{equation}

The other requirement is to reach the goal by executing an ideal or expected behavior. We can now enforce the constraints (equations \ref{c1:eq} - \ref{c3:eq}), by introducing a new reward objective to drive the action errors to zero. Equation \ref{rew2:eq} defines the action error reward given the extended state (equation \ref{new_state:eq}). Now, it is imperative to rightly define these action errors. A straightforward way would be to simply have these errors as the difference between the next observed state and the expected next state from the model description (equation \ref{state_trans:eq}), i.e. set $x_e = x_{t+1} - (x_{t} + \Delta x), y_e = y_{t+1} - (y_{t} + \Delta y), z_e = z_{t+1} - (z_{t} + \Delta z)$. However, the RL agent can easily exploit the second reward objective (equation \ref{rew1:eq}) by taking a small step to reset the reward to 0 and then followed by a longer inconsistent stride. To circumvent this, we can introduce a cumulative error term in the equation (equation \ref{err_states:eq}), where $\alpha < 1$ is a discount factor and $x_e'$ is the previous error (from the previous timestep). The discount factor was observed to be crucial, which would otherwise induce too many oscillations in the control analogous to pure integral control. Similarly, we define the error states $y_e, z_e$.

\begin{align}
	\label{err_states:eq}
	x_e = \Sigma_{t'=0}^t \alpha^{t - t'} (x_{t+1} - (x_t + \Delta x_t)) \notag \\
	= \alpha  (x_{t+1} - (x_t + \Delta x_t)) + \alpha x_e'
\end{align}

The reward function from equation \ref{rew2:eq} resembles a typical LQR cost function, however, there are stark differences when looked at closely. Consider the partial state of the system encapsulating the position of the quadrotor in space, $p_t = [x_t, \ y_t, \ z_t]^T$. Assuming the dynamics to be of the form $p_{t+1} = \Phi p_t + \Gamma u_t$, a typical LQR cost would look like $\Sigma_t p_t^T Q_1 p_t + u_t^T Q_2 u_t + p_t Q_{12} u_t$. In the case of a stochastic LQR problem, the dynamics would often evolve as $p_{t+1} = \Phi p_t + \Gamma u_t + G w_t$, where $w_t$ is usually assumed to be a Gaussian noise with zero mean, which would yield the same expected cost function as in the deterministic LQR case. However, the point to note here is that the standard LQR cost penalizes the magnitude of the control signal $u_t$, while the RL objective function proposed in equation \ref{rew2:eq} does not, instead, the penalty would be proportional to the deviation from the expected next state $x_t + u_t$ defined by the dynamics. The controller is free to choose an arbitrarily large control signal as long as it can execute it successfully, this contrasts a standard LQR control approach. This will be evident from experiments carried out in the subsequent sections. The same argument can also be extended to other error states $y_e, z_e$.

Given that $\mathbb{E} [p_{t+1}] = \Phi p_t + \Gamma u_t$ (since  $\mathbb{E} [w_t] = 0$), one might wonder if it would be redundant to optimize the agent to counteract the disturbance in the first place. In this context let $e_t = [x_e, \ y_e, \ z_e]$ be the deviation of the agent from its ideal state after executing the action $u_t$ from state $p_t$ as a consequence of $w_t$ (external wind, actuator noise, noise sensor data, etc). The net effect of $e_t$ can thus be measured as the Euclidean distance from its ideal next state $p_t + u_t$ (equation \ref{dist_eff_err:eq}.

\begin{equation}
	\label{dist_eff_err:eq}
	e_t = \sqrt{x_e^2 + y_e^2 + z_e^2}
\end{equation}

\begin{figure*}[h]
	\centering
	\caption{Random Walk Experiment}
	\begin{subfigure}[b]{0.25\textwidth}
		\caption{Trajectory}
		\includegraphics[width=\textwidth]{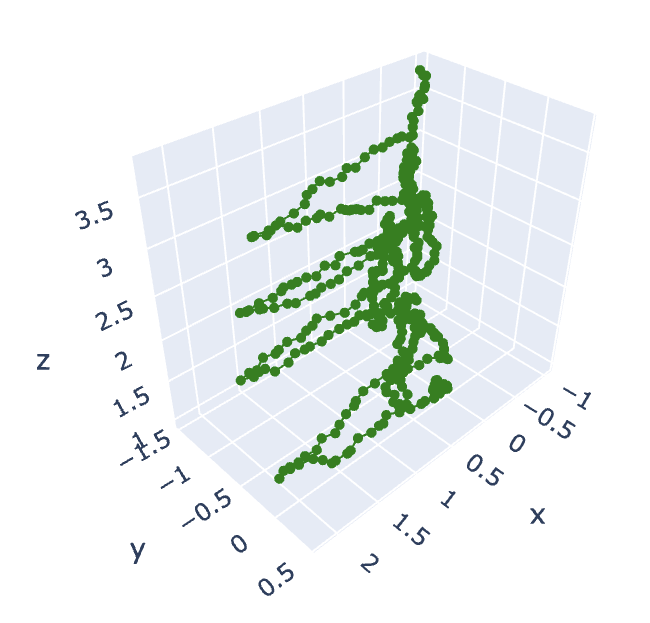}
		\label{r_walk_traj:fig}
	\end{subfigure}
	\hfill
	\begin{subfigure}[b]{0.45\textwidth}
		\caption{Error Distribution}
		\includegraphics[width=\textwidth]{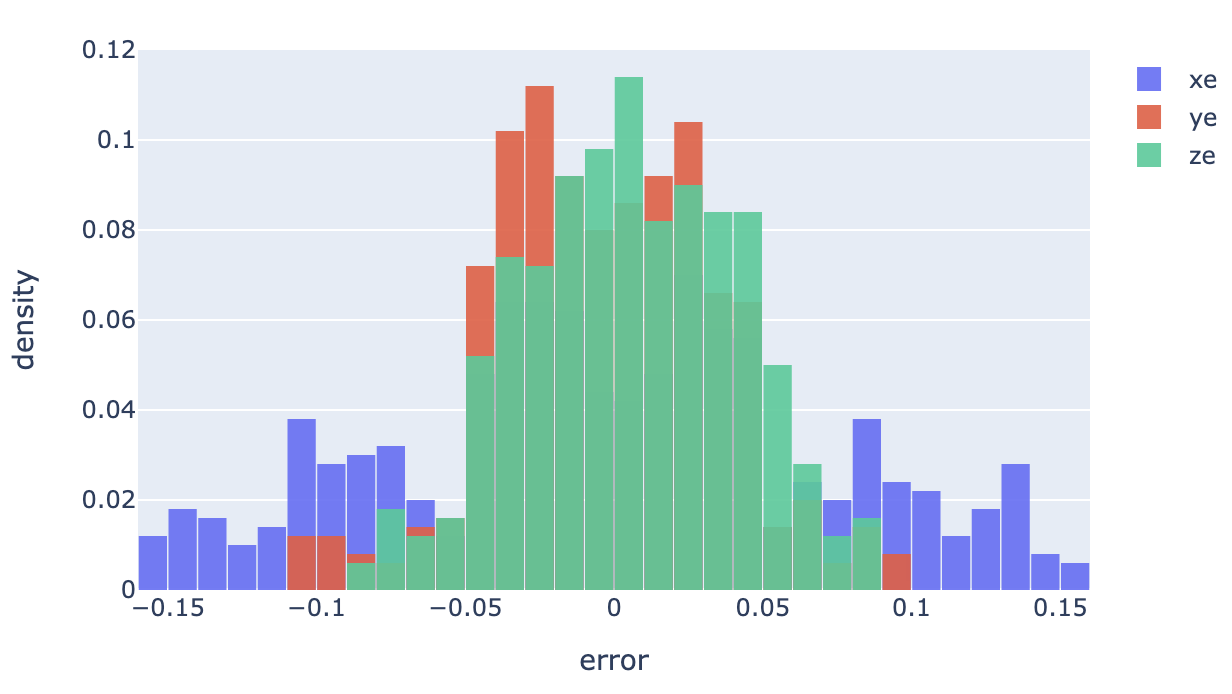}
		\label{r_walk_err_dist:fig}
	\end{subfigure}
	\hfill
	\begin{subfigure}[b]{0.25\textwidth}
		\caption{Error Magnitude ($|\text{action}|_2 - |\text{executed}|_2$, in meters, the lower the better)}
		\includegraphics[width=\textwidth]{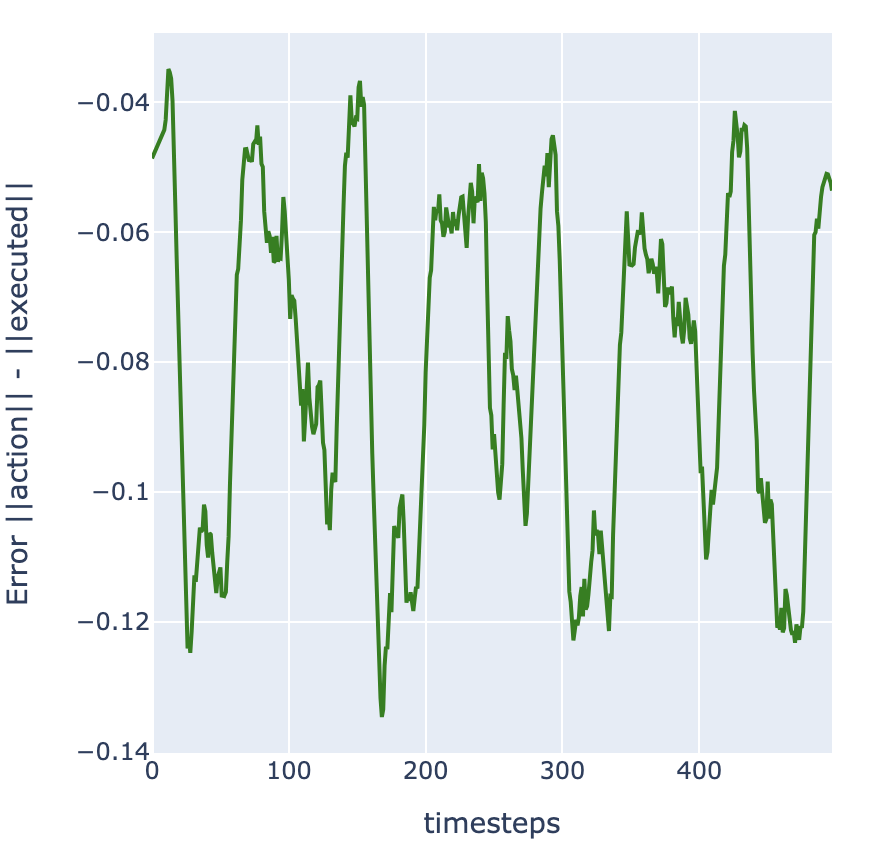}
		\label{r_walk_err_mag:fig}
	\end{subfigure}
\end{figure*}

To analyze the characteristics of the disturbances acting on the system, a random walk was simulated by sampling the high-level actions uniformly as per equation \ref{a_limits:eq} serving as relative coordinates for the quadrotor to move to and the low-level PID controller executing the actions while subjecting the quadrotor to random step disturbances (figure \ref{train_dists:fig}). The resulting error distributions are shown in figure \ref{r_walk_err_dist:fig}. The individual distributions of $x_e, y_e, z_e$ are normally distributed with zero means (figure \ref{r_walk_err_dist:fig}). The means were estimated as $\mathbb{E}[x_e] = \frac{1}{N} \ \Sigma_i x_{e_i}$ and the covariances as $\mathbb{C}[x_e, y_e] = \frac{1}{N} \ \Sigma_i x_{e_i} y_{e_j} - \mathbb{E}[x_e] \mathbb{E}[y_e]$ (similarly $\mathbb{C}[x_e, z_e], \mathbb{C}[y_e, z_e]$), table \ref{r_walk_stats:table}. It can be observed that although the means are close to zero, the covariances seem quite high to ignore. The error $e_t$ can now be represented by a multivariate normal distribution using the estimated means and covariances (table \ref{r_walk_stats:table}) and the expected value of equation \ref{dist_eff_err:eq} can be calculated by sampling the probability density function and approximating the expectation by averaging under the law of large numbers (equation \ref{r_walk_mag_est:eq}). The resulting estimate was found to be $0.08168$ (meters), i.e. the quadrotor is was found to be off by a magnitude of 8 cm from its ideal next state. This can also be verified by inspecting figure \ref{r_walk_err_mag:fig}, which shows the magnitude of the error (the difference between the magnitudes of predicted action and the executed maneuver), and the mean of the distribution exactly matches the estimated deviation of 8 cm. It can also be observed that the error can even grow as large as a staggering 13 cm per step, which is well above the specified action limits (equation \ref{a_limits:eq}). Thus, the conclusion is that although the error distributions can be modeled as a zero mean noise acting on the system, it would be unwise to not compensate for them, at least in this particular task.

\begin{table}[h]
\centering
\caption{Error Statistics}
\label{r_walk_stats:table}
\begin{tabular}{|c|c|c|c|c|}
\hline
\textbf{} & \multicolumn{3}{c|}{\textbf{Covariances}} & \textbf{Means} \\ \hline
          & $x_e$        & $y_e$        & $z_e$       &                \\ \hline
$x_e$     & 0.07218      & 0.02731      & 0.01915     & $-0.001759$    \\ \hline
$y_e$     & 0.02731      & 0.03799      & 0.01650     & $-0.002713$    \\ \hline
$z_e$     & 0.01915      & 0.01650      & 0.03614     & $0.0055015$    \\ \hline
\end{tabular}
\end{table}

\begin{align}
\label{r_walk_mag_est:eq}
\mathbb{E}\left[e_{t}\right] & =\int_{e_t} e_t \ P(e_t) \ d e_{t} \notag \\
& =\int_{x_{e}} \int_{y_{e}} \int_{z_{e}} \sqrt{x_{e}^{2}+y_{e}^{2}+z_{e}^{2}} \ P(x_e, y_e, z_e) \ d x_{e} \ d y_{e} \ d z_{e} \notag \\
& \approx \lim_{N \to \infty} \frac{1}{N} \sum_{i} \sqrt{x_{e}^{2}+y_{e}^{2}+z_{e}^{2}} \ \ , \ \left[\begin{array}{l}
x_{e} \\
y_{e} \\
z_{e}
\end{array}\right] \sim \mathcal{N}(\mu_e, \Sigma_e)
\end{align}

\section{Single-objective Reinforcement Learning}

The goal here is to learn a strategy to navigate the quadrotor to a fixed point in space. The quadrotor starts off at an initial state $s_0$ and learns to navigate to the point $[0, 0, 1]$ during the training phase. To test the theory and claims made in the previous sections, a rigorous comparison is made on the following approaches:
\begin{enumerate*}[label=(\roman*)]
  \item Baseline model: This is a vanilla reinforcement learning model formulated for the navigation task. It uses the kinematics state space (equation \ref{obs_space:eq}). The reward function only consists of the navigation reward (equation \ref{rew1:eq}) and the agent is not subjected to any form of disturbances during the training phase.
  \item Baseline model with disturbances: The formulation is very similar to the baseline model (i), except that the model experiences disturbances during the training phase.
  \item Model with error correction (proposed approach for robustness): This is a model with disturbances (ii) but uses the extended state space (equation \ref{err_states:eq}) and also receives a feedback regarding the state errors which was introduced earlier(equation \ref{rew2:eq}). Since the model sees two distinct reward objectives, the expected scalarized return (ESR) approach is employed by defining a utility function to scalarize the rewards prior to the expectation (equation \ref{utility_fun:eq}).
  \item An LSTM model: The problem setting is exactly similar to (iii) except that the reinforcement learning agent has an LSTM architecture. Having the LSTM architecture dissents from the rest of the options adhering to the MDP assumptions, the motivation for including the LSTM architecture in the comparison was to explore the possibility of the RL agent learning to behave robustly given a sequence of previous states, i.e. whether a pure deep learning architecture can render the proposed method obsolete.
\end{enumerate*}
To draw a fair comparison, all of the above approaches use the exact same underlying reinforcement learning algorithm, Proximal Policy Gradient (PPO) \cite{schulman2017proximal}. PPO is a stochastic policy gradient algorithm that improves on the prior policy gradient implementations like Trusted Region Policy Optimization \cite{schulman2017trust}. The algorithmic implementations are made available by stable-baselines3 \cite{stable-baselines3}.

\begin{equation}
	\label{utility_fun:eq}
	U \left(
	\begin{bmatrix}
	R_{nav} \\
	R_{err}
	\end{bmatrix}
	\right) = w^T \begin{bmatrix}
	R_{nav} \\
	R_{err}
	\end{bmatrix} = \left[1 \ \ 0.5 \right] \begin{bmatrix}
	R_{nav} \\
	R_{err}
	\end{bmatrix}
\end{equation}

\begin{figure}[h]
	\centering
	\caption{Training Disturbances}
	{$X, Y, Z$ disturbances applied during training. The disturbances are flipped every 20 simulation steps.}
	\includegraphics[scale=0.25]{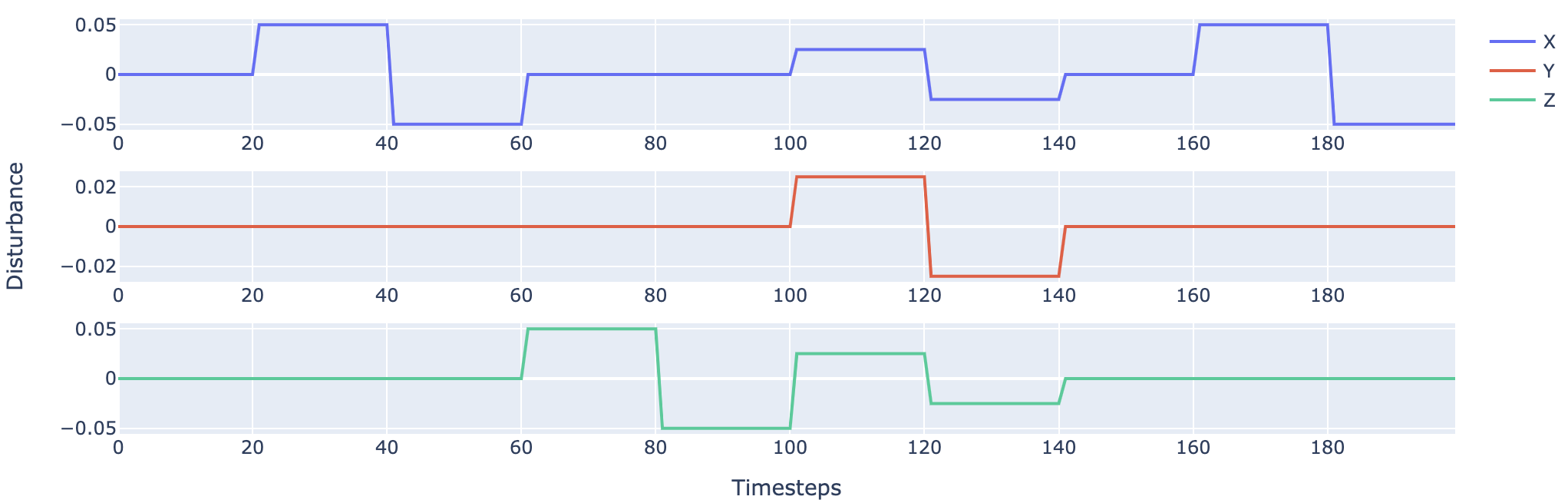}
	\label{train_dists:fig}
\end{figure}

The disturbances during the training phase are feeble step/pulse disturbances acting along one or all of the $X, Y, Z$ directions emulating a wind force. Figure \ref{train_dists:fig} shows the disturbances applied during the training phase, the magnitude and/or the direction of the disturbances change every 20 simulation steps, with the intention of simulating real-world disturbances that are not completely random and for the RL agent to make ad-hoc adjustments to adapt to the disturbances. During the evaluation phase, the magnitude and the direction of the step disturbance remain fixed throughout and the performance is evaluated against disturbances of varying magnitudes with much stronger intensities. The initial state is set to $[0, \ 0, \ 0]$ at the start of every training episode. Although the RL agent would benefit from random initialization, the intention is to keep the comparison fair for all the approaches by eliminating the randomness in the environment. However, the initial state is offset during the evaluation phase to look at the agent's behavior to work with and against the disturbances acting on it. Table \ref{train_dists:fig} lists the training hyperparameters.

\begin{table}[H]
\caption{Training Hyperparameters}
	\label{training_hyperparams:table}
	\centering
\begin{tabular}{|l|l|}
\hline
\textbf{Hyperparameter}                        & \textbf{Value}      \\ \hline
Episode Length (in secs)                       & 2                   \\ \hline
Initial XYZ state                              & $[0, \ 0, \ 0]$     \\ \hline
Disturbance $|X_{max}|$                        & 0.05                \\ \hline
Disturbance $|Z_{max}|$                        & 0.05                \\ \hline
Disturbance $|XYZ_{max}|$                      & 0.025               \\ \hline
Disturbance Flip Frequency                     & 20 simulation steps \\ \hline
Number of Timesteps                            & 1,000,000             \\ \hline
Neural Net Policy Architecture (hidden layers) & $[64, \ 64]$        \\ \hline
\end{tabular}
\end{table}

Multiple evaluation metrics are formulated to effectively assess and compare the pros and cons of the different approaches described above. We obviously have the negative of squared reward and a few other metrics to evaluate the robustness. The distance traveled $\Sigma_t |p_{t} - p_{p+1}|_2$ is measured as the physical distance traveled by the agent while executing the trajectory as opposed to the typical cumulative RL reward. Here $p_t$ denotes the state comprising of the XYZ coordinates of the quadrotor $p_T = [x_t, \ y_t, \ z_t]$. The smoothness of the trajectory is measured as the mean of the second derivative, i.e. the rate of change of direction of the quadrotor, $\frac{1}{N}\int |\tau''(t)|_2 dt$. Ideally, we would desire the system to navigate in a smooth trajectory, as opposed to a zig-zag or jagged trajectory. As the ascent phase of the quadrotor is crucial prior to it reaching the destination and start hovering, the average ascent step is computed during the initial few timesteps of the trajectory. A plot of intended minus the executed distance ($|u_t|_2 - |p_t - p_{t+1}|_2$) is tracked, ideally, we would like this to be zero, i.e. the quadrotor should be able to execute the actions successfully. Finally, the ``converged'' metric indicates whether the quadrotor managed to reach the destination within a 10 cm accuracy. 

\begin{figure*}[h]
	\centering
	\caption{Evaluation Rewards. \newline Models compared: i. \textbf{lstm}: an LSTM agent. ii. \textbf{dist-err}: the proposed arch for single-objective, receives feedback for the action errors. iii. \textbf{dist-err-u}: similar to LQR cost, agent receives a negative L2 reward of the control/action taken. iv. \textbf{dist}: baseline model subjected to disturbances during training. v. \textbf{baseline}: the baseline model.}
	\begin{subfigure}[b]{0.45\textwidth}
		\caption{Disturbance along X}
		\includegraphics[width=\textwidth]{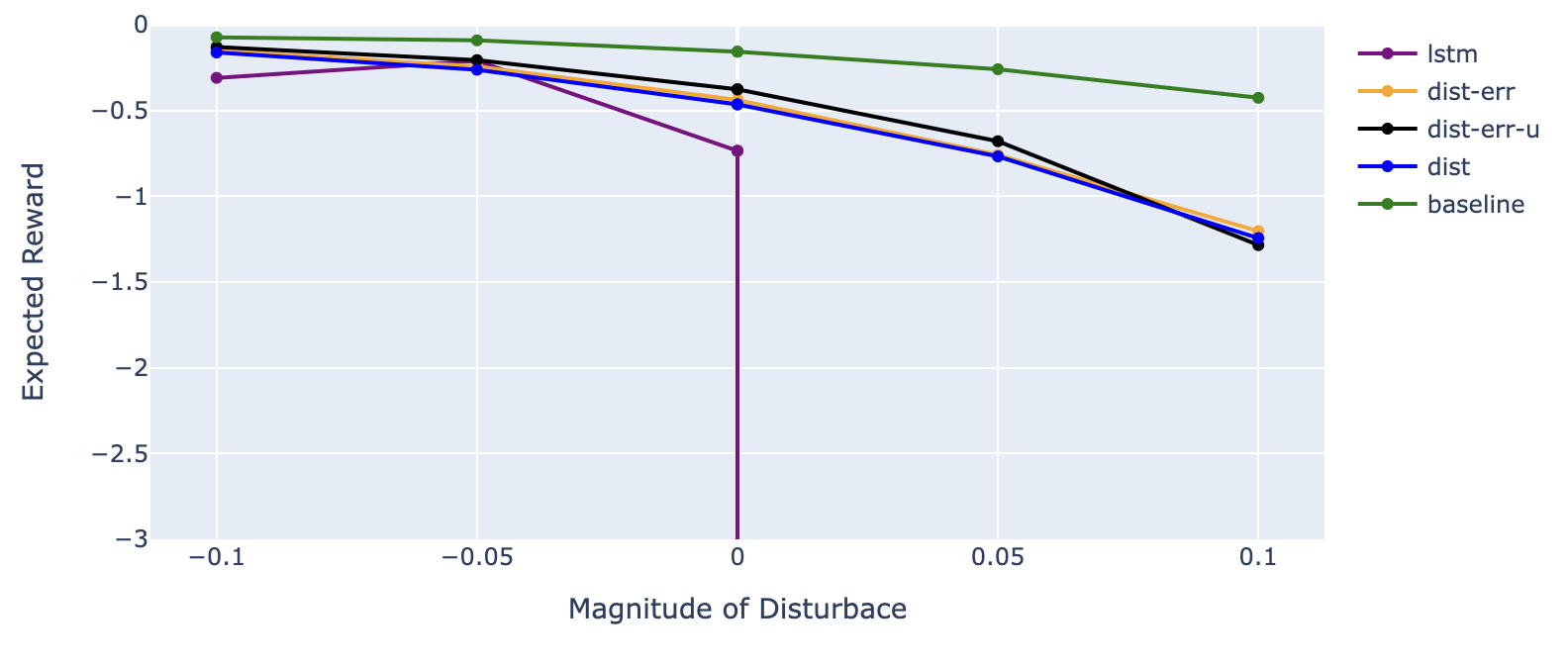}
		\label{rew_x:fig}
	\end{subfigure}
	\hfill
	\begin{subfigure}[b]{0.45\textwidth}
		\caption{Disturbance along Z}
		\includegraphics[width=\textwidth]{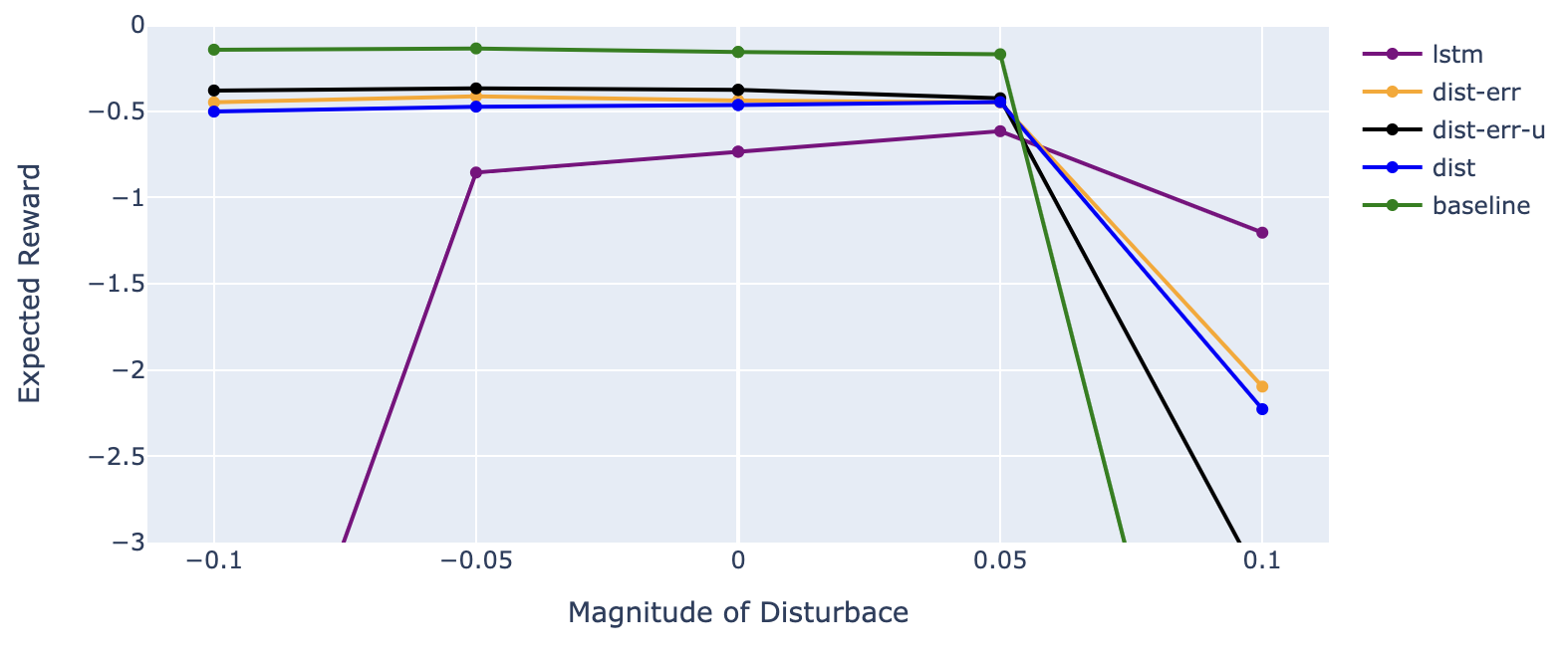}
		\label{rew_z:fig}
	\end{subfigure}
	\hfill
	\begin{subfigure}[b]{0.45\textwidth}
		\caption{Disturbance along XYZ}
		\includegraphics[width=\textwidth]{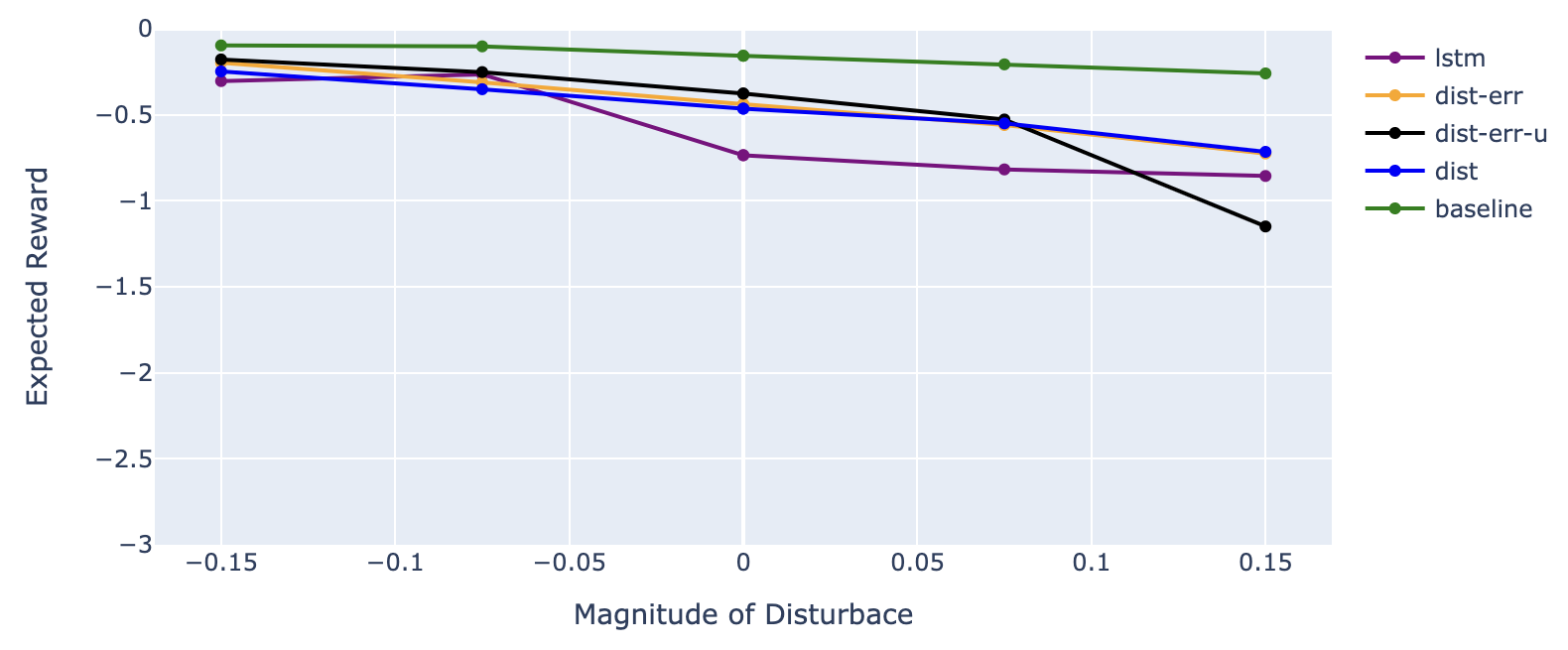}
		\label{rew_xyz:fig}
	\end{subfigure}
\end{figure*}

\begin{figure*}[h]
	\centering
	\caption{Evaluation Trajectories and Action Errors During Ascent. \newline The figures on top plot the trajectories of the agent during the evaluation phase. The figures at the bottom compare the action errors $|\text{action}|_2 - |\text{executed}|_2$ (in meters, the lower the better). \newline Models compared: i. \textbf{lstm}: an LSTM agent. ii. \textbf{dist-err}: the proposed arch for single-objective, receives feedback for the action errors. iii. \textbf{dist-err-u}: similar to LQR cost, agent receives a negative L2 reward of the control/action taken. iv. \textbf{dist}: baseline model subjected to disturbances during training. v. \textbf{baseline}: the baseline model. }
	\begin{subfigure}[b]{0.32\textwidth}
		\caption{Disturbance: -0.050 along X}
		\includegraphics[width=\textwidth]{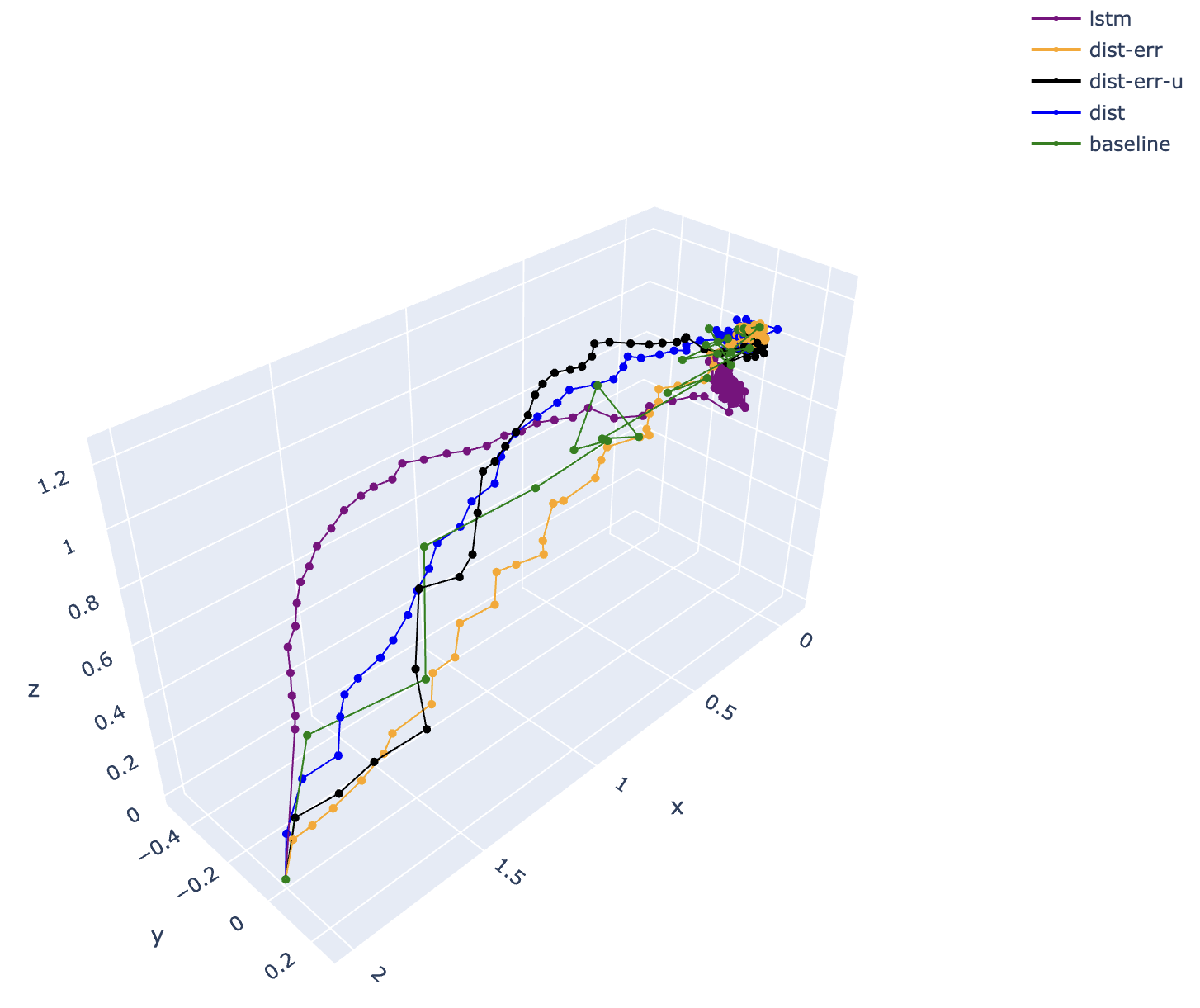}
		\label{t_x:fig}
	\end{subfigure}
	\hfill
	\begin{subfigure}[b]{0.33\textwidth}
		\caption{Disturbance: +0.050 along Z}
		\includegraphics[width=\textwidth]{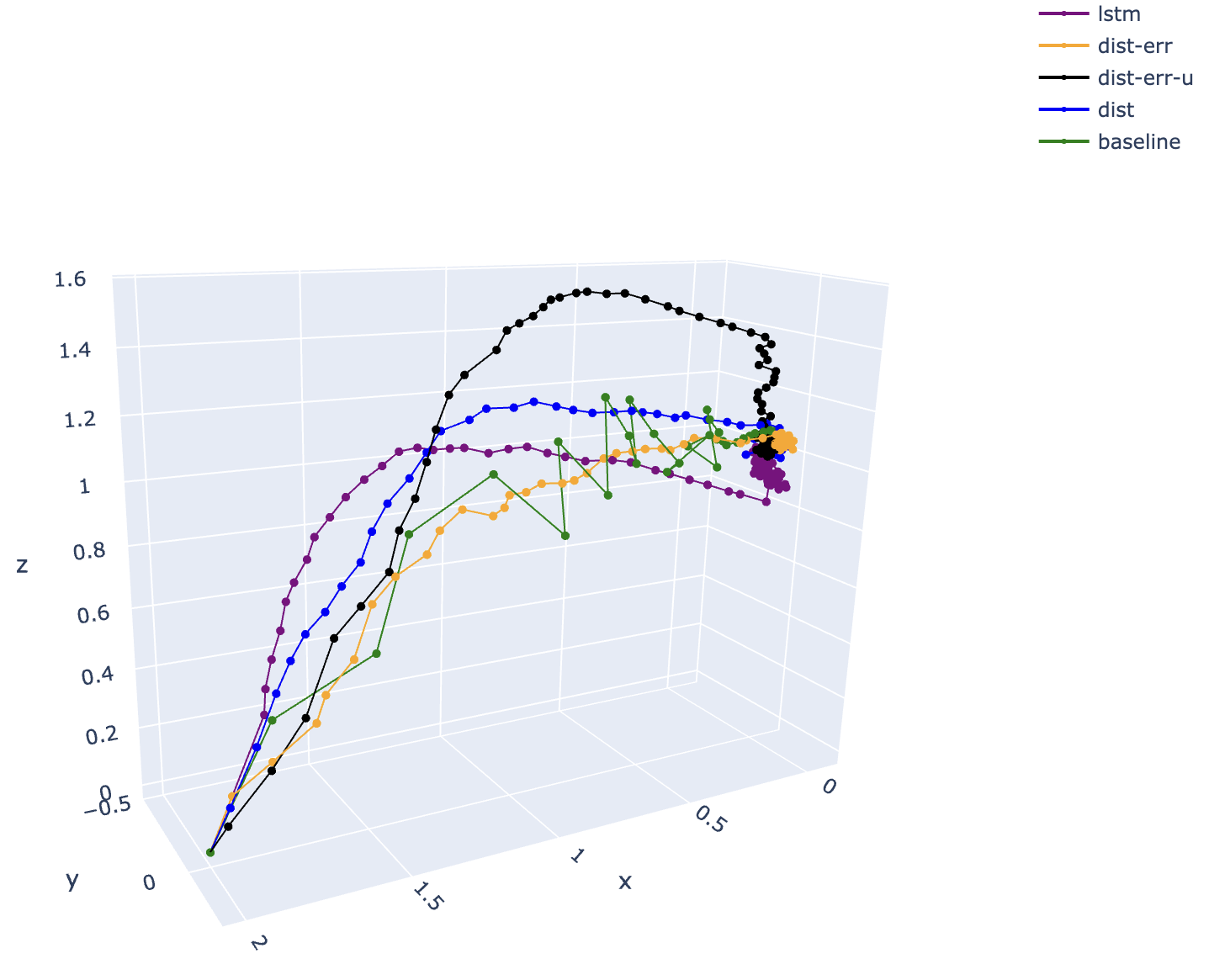}
		\label{t_z:fig}
	\end{subfigure}
	\hfill
	\begin{subfigure}[b]{0.32\textwidth}
		\caption{Disturbance: +0.075 along XYZ}
		\includegraphics[width=\textwidth]{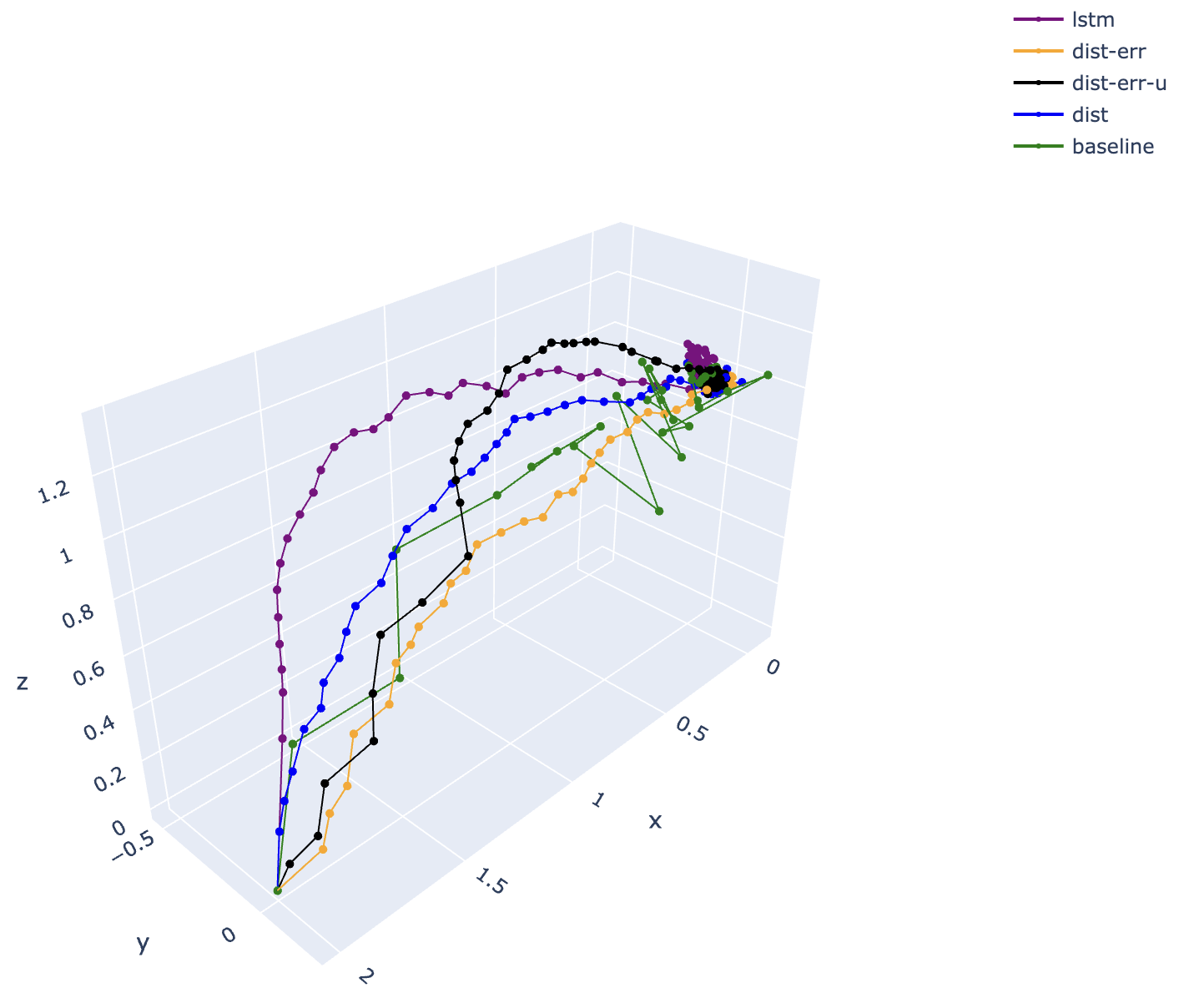}
		\label{t_xyz:fig}
	\end{subfigure}
	\hfill
	\begin{subfigure}[b]{0.3\textwidth}
		\caption{Disturbance: -0.050 along X}
		\includegraphics[width=\textwidth]{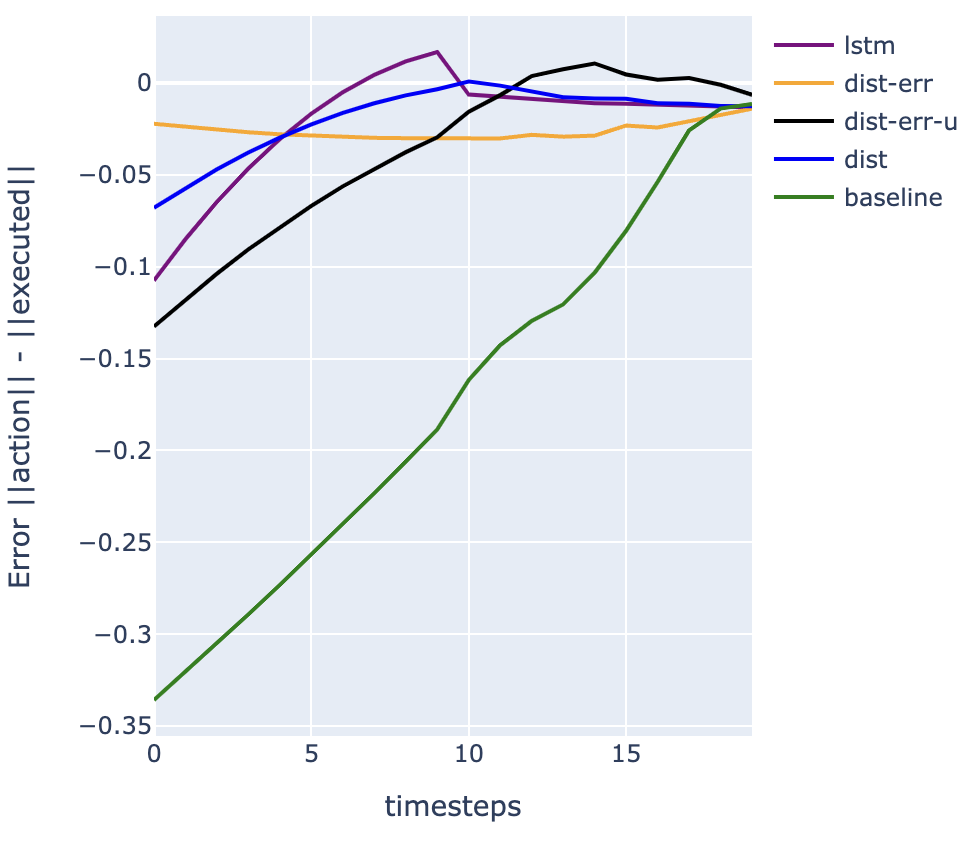}
		\label{ae_x:fig}
	\end{subfigure}
	\hfill
	\begin{subfigure}[b]{0.3\textwidth}
		\caption{Disturbance: +0.050 along Z}
		\includegraphics[width=\textwidth]{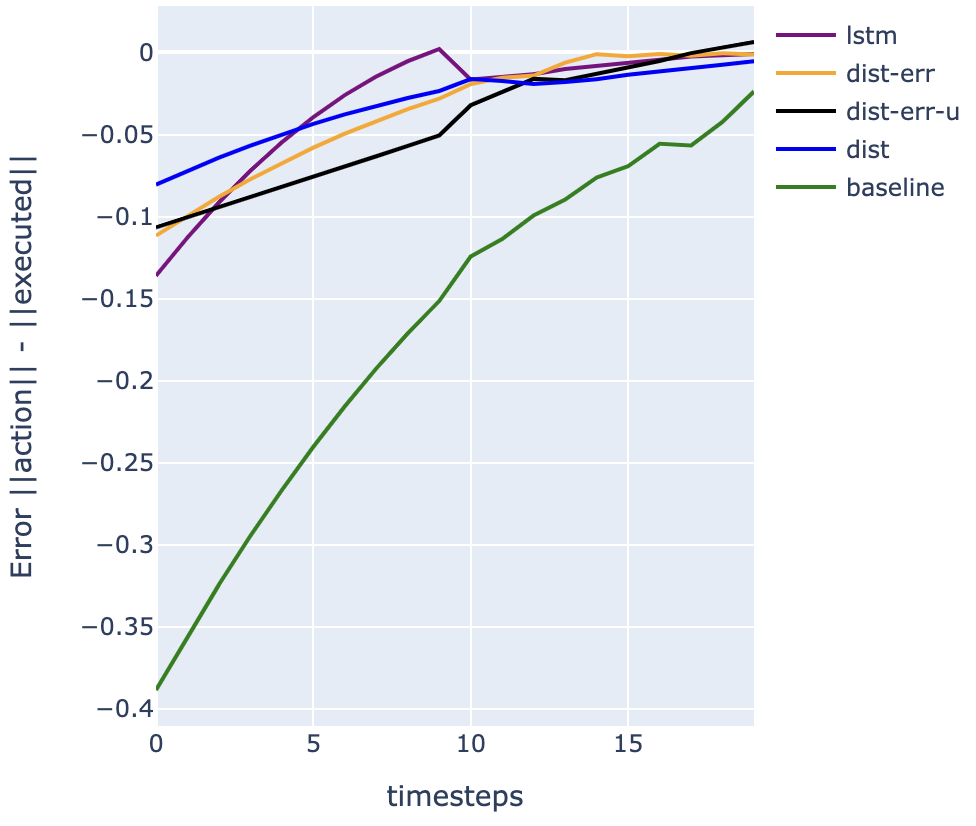}
		\label{ae_z:fig}
	\end{subfigure}
	\hfill
	\begin{subfigure}[b]{0.3\textwidth}
		\caption{Disturbance: +0.075 along XYZ}
		\includegraphics[width=\textwidth]{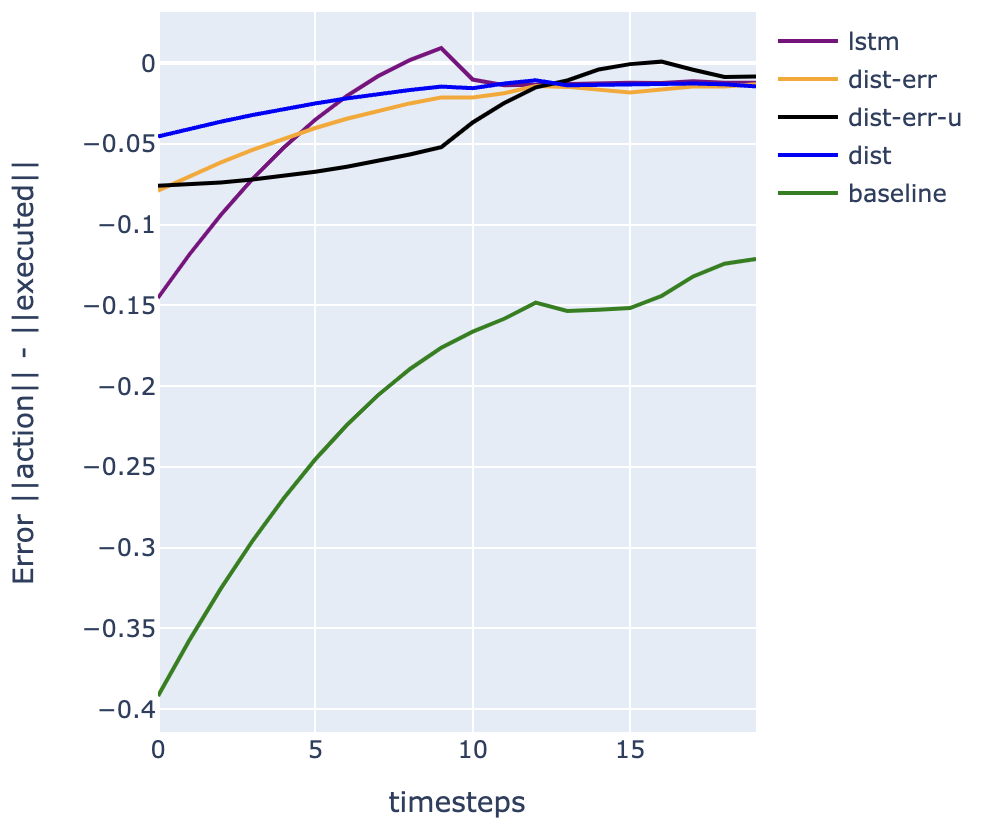}
		\label{ae_xyz:fig}
	\end{subfigure}
\end{figure*}

Figures \ref{rew_x:fig} - \ref{rew_xyz:fig} plot the evaluated reward functions for different combinations and magnitudes of XYZ disturbances. The evaluations were captured by setting the disturbances to be constant throughout the episode, unlike the training phase where the disturbances change directions randomly. The curves are asymmetric due to the fact that the initial state of the drone is moved to $[2, \ 0, \ 0]$ for evaluation. So, the drone is assisted by the wind when the disturbance is along the negative X direction and opposed when positive. The first thing to notice is that the baseline model, which was not exposed to any form of disturbance during the training phase outperforms the rest of the pack by a considerable margin in terms of the cumulative RL reward, albeit the other models achieved a higher reward in a few extreme cases (figure \ref{rew_z:fig}). This finding corroborates the claims made by \cite{glossop2022characterising}. However, the reward functions do not paint the complete picture. Figures \ref{t_x:fig} - \ref{t_xyz:fig} plot the evaluation trajectories of policies for different cases of evaluation. Looking closely at the trajectories we find that the baseline model takes longer strides and manages to reach the in fewer RL steps, while the rest of the pack takes a considerably more number of RL steps during ascent. This explains the reason why the baseline model achieves the best RL reward, as the RL reward is a cumulative discounted sum of the reward, the more RL steps the policy executes, the more negative reward is accumulated. While achieving a high RL reward is always the goal in reinforcement learning, it can also have adverse consequences. The figures \ref{ae_x:fig} - \ref{ae_xyz:fig} spotlight the underlying problem, the baseline model appears to be a true outlier. As per the design formulation \ref{a_limits:eq}, the maximum possible relative distance the agent was supposed to execute was $\sqrt{0.05^2 + 0.05^2 + 0.05^2} = 0.0866 \ \text{m or 8.66 cm}$. However, the baseline model (green curve) seems to be encountering an enormous amount of action error, starting off from a staggering 30 cm and then gradually decreasing to 0 cm after the ascent period. An error of 30 cm essentially means that the quadrotor was off by 30 cm from where it planned to go by executing the particular action. In contrast, the models trained with disturbances (the rest of the pack) appear to be much more stable, and the proposed model (``dist-err'') in particular appears to do a better job, implying that the agent was able to execute the actions with sufficient certainty. This further supports the assumptions made by simulating the random walk experiment earlier, that ignoring the effects of disturbances by assuming the normal nature of the errors (figure \ref{r_walk_err_dist:fig}) would not be the best thing to do, because although the expected values of errors are zero, the effect on the quadrotor is actually $\mathbb{E}[\sqrt{x_e^2 + y_e^2 + z_e^2}]$ which was found to be much larger and not equal to zero. Table \ref{metrics:table} further lists more insightful metrics captured and spotlights the proposed method. Although the ``baseline'' model outperformed in terms of the RL reward, surprisingly it does not hold up in terms of the actual physical distance traveled by the quadrotor. ``dist-err'' manages to reach the destination of $[0, \ 0, \ 1]$ from $[2, \ 0, \ 0]$ by traveling a significantly shorter distance compared to the baseline. The rest of the models perform similarly, but ``dist-err'' appears to have an edge in most cases. This can also be viewed by inspecting the trajectory plots (figures \ref{t_x:fig} - \ref{t_xyz:fig}), where the trajectory appears to be much more focused for ``dist-err''. The ``dist-err'' model also appears to be much smoother, this is evident from the fact that the mean of the magnitude of the second derivate is much smaller (group 2, table \ref{metrics:table}). All the models trained with disturbance seem to perform similarly when it comes to the average ascent step, albeit ``dist-err-u'' with a slight edge. However, the ``baseline'' again appears to be an outlier, taking significantly longer strides during the ascent phase. It can also be observed that the  ``dist-err'' model appears to converge to the goal under all the experiments except the last one with a disturbance of 0.100 acting along a positive Z direction. While the rest of the models fail to reach the goal under multiple scenarios.

The conclusion we draw from these results is that a higher RL reward does not necessarily imply the inherent robustness of the models to all aspects of evaluation. There could be severe consequences or adverse effects surrounding the process dynamics or other relevant metrics to look at to truly assess the robustness. Thus, enforcing the dynamics by a reward function (model-assisted) learning appears to have some positive effect with regards to the certainty of RL algorithms.

\begin{landscape}
\begin{table}[]
\caption{Single-objective Navigation Metrics. \newline Models compared: i. \textbf{lstm}: an LSTM agent. ii. \textbf{dist-err}: the proposed arch for single-objective, receives feedback for the action errors. iii. \textbf{dist-err-u}: similar to LQR cost, agent receives a negative L2 reward of the control/action taken. iv. \textbf{dist}: baseline model subjected to disturbances during training. v. \textbf{baseline}: the baseline model.}
\label{metrics:table}
\centering
\begin{adjustbox}{height=0.28\textheight}
\def\arraystretch{2.5}
\begin{tabular}{|c|c|ccccc|ccccc|ccccc|ccccc|}
\hline
\textbf{}          & \textbf{}           & \multicolumn{5}{c|}{\textbf{Distance Travelled}}                                                                                                                                 & \multicolumn{5}{c|}{\textbf{Smoothness}}                                                                                                                                            & \multicolumn{5}{c|}{\textbf{Average Ascent Step (L2 Norm of Action)}}                                                                                                                   & \multicolumn{5}{c|}{\textbf{Converged}}                                                                                                                                         \\ \hline
\textbf{Direction} & \textbf{Mag. Dist.} & \multicolumn{1}{c|}{\textbf{lstm}} & \multicolumn{1}{c|}{\textbf{dist-err}}  & \multicolumn{1}{c|}{\textbf{dist-err-u}} & \multicolumn{1}{c|}{\textbf{dist}} & \textbf{baseline} & \multicolumn{1}{c|}{\textbf{lstm}} & \multicolumn{1}{c|}{\textbf{dist-err}} & \multicolumn{1}{c|}{\textbf{dist-err-u}} & \multicolumn{1}{c|}{\textbf{dist}}     & \textbf{baseline} & \multicolumn{1}{c|}{\textbf{lstm}}     & \multicolumn{1}{c|}{\textbf{dist-err}} & \multicolumn{1}{c|}{\textbf{dist-err-u}} & \multicolumn{1}{c|}{\textbf{dist}}     & \textbf{baseline} & \multicolumn{1}{c|}{\textbf{lstm}} & \multicolumn{1}{c|}{\textbf{dist-err}} & \multicolumn{1}{c|}{\textbf{dist-err-u}} & \multicolumn{1}{c|}{\textbf{dist}} & \textbf{baseline} \\ \hline
x                  & -0.050              & \multicolumn{1}{c|}{6.599416}      & \multicolumn{1}{c|}{6.132900}           & \multicolumn{1}{c|}{\textbf{5.714949}}   & \multicolumn{1}{c|}{6.111227}      & 6.618811          & \multicolumn{1}{c|}{0.009211}      & \multicolumn{1}{c|}{0.008646}          & \multicolumn{1}{c|}{\textbf{0.006601}}   & \multicolumn{1}{c|}{0.007670}          & 0.032359          & \multicolumn{1}{c|}{0.138675}          & \multicolumn{1}{c|}{\textbf{0.099394}} & \multicolumn{1}{c|}{0.173148}            & \multicolumn{1}{c|}{0.123149}          & 0.383917          & \multicolumn{1}{c|}{Yes}           & \multicolumn{1}{c|}{Yes}               & \multicolumn{1}{c|}{Yes}                 & \multicolumn{1}{c|}{Yes}           & Yes               \\ \hline
x                  & -0.100              & \multicolumn{1}{c|}{6.499861}      & \multicolumn{1}{c|}{\textbf{5.331071}}  & \multicolumn{1}{c|}{5.918821}            & \multicolumn{1}{c|}{5.719601}      & 5.798424          & \multicolumn{1}{c|}{0.007364}      & \multicolumn{1}{c|}{0.005405}          & \multicolumn{1}{c|}{0.006002}            & \multicolumn{1}{c|}{0.005149}          & 0.022950          & \multicolumn{1}{c|}{0.156945}          & \multicolumn{1}{c|}{0.114534}          & \multicolumn{1}{c|}{0.168128}            & \multicolumn{1}{c|}{\textbf{0.112654}} & 0.411196          & \multicolumn{1}{c|}{Yes}           & \multicolumn{1}{c|}{Yes}               & \multicolumn{1}{c|}{Yes}                 & \multicolumn{1}{c|}{Yes}           & Yes               \\ \hline
x                  & 0.000               & \multicolumn{1}{c|}{7.603805}      & \multicolumn{1}{c|}{5.364380}           & \multicolumn{1}{c|}{\textbf{5.280013}}   & \multicolumn{1}{c|}{6.599345}      & 5.843086          & \multicolumn{1}{c|}{0.011430}      & \multicolumn{1}{c|}{0.005480}          & \multicolumn{1}{c|}{\textbf{0.005477}}   & \multicolumn{1}{c|}{0.011106}          & 0.029162          & \multicolumn{1}{c|}{0.155782}          & \multicolumn{1}{c|}{0.129354}          & \multicolumn{1}{c|}{0.133528}            & \multicolumn{1}{c|}{\textbf{0.099125}} & 0.380344          & \multicolumn{1}{c|}{Yes}           & \multicolumn{1}{c|}{Yes}               & \multicolumn{1}{c|}{Yes}                 & \multicolumn{1}{c|}{Yes}           & Yes               \\ \hline
x                  & 0.050               & \multicolumn{1}{c|}{5.256427}      & \multicolumn{1}{c|}{\textbf{4.934614}}  & \multicolumn{1}{c|}{5.306519}            & \multicolumn{1}{c|}{7.602165}      & 6.909383          & \multicolumn{1}{c|}{0.040803}      & \multicolumn{1}{c|}{\textbf{0.003857}} & \multicolumn{1}{c|}{0.004457}            & \multicolumn{1}{c|}{0.015453}          & 0.036542          & \multicolumn{1}{c|}{0.968272}          & \multicolumn{1}{c|}{0.099889}          & \multicolumn{1}{c|}{0.164785}            & \multicolumn{1}{c|}{\textbf{0.099618}} & 0.378084          & \multicolumn{1}{c|}{\textbf{No}}   & \multicolumn{1}{c|}{Yes}               & \multicolumn{1}{c|}{Yes}                 & \multicolumn{1}{c|}{Yes}           & Yes               \\ \hline
x                  & 0.100               & \multicolumn{1}{c|}{8.016002}      & \multicolumn{1}{c|}{5.009077}           & \multicolumn{1}{c|}{5.364408}            & \multicolumn{1}{c|}{7.573557}      & \textbf{4.562196} & \multicolumn{1}{c|}{0.232892}      & \multicolumn{1}{c|}{\textbf{0.003696}} & \multicolumn{1}{c|}{0.004433}            & \multicolumn{1}{c|}{0.015047}          & 0.033325          & \multicolumn{1}{c|}{1.349194}          & \multicolumn{1}{c|}{0.125718}          & \multicolumn{1}{c|}{0.173467}            & \multicolumn{1}{c|}{\textbf{0.120015}} & 0.599928          & \multicolumn{1}{c|}{\textbf{No}}   & \multicolumn{1}{c|}{Yes}               & \multicolumn{1}{c|}{Yes}                 & \multicolumn{1}{c|}{Yes}           & \textbf{No}       \\ \hline
xyz                & -0.075              & \multicolumn{1}{c|}{6.836447}      & \multicolumn{1}{c|}{\textbf{5.335730}}  & \multicolumn{1}{c|}{5.959134}            & \multicolumn{1}{c|}{6.811069}      & 6.524232          & \multicolumn{1}{c|}{0.010284}      & \multicolumn{1}{c|}{\textbf{0.005464}} & \multicolumn{1}{c|}{0.007842}            & \multicolumn{1}{c|}{0.012476}          & 0.034549          & \multicolumn{1}{c|}{0.131541}          & \multicolumn{1}{c|}{\textbf{0.111868}} & \multicolumn{1}{c|}{0.127855}            & \multicolumn{1}{c|}{0.116927}          & 0.384197          & \multicolumn{1}{c|}{Yes}           & \multicolumn{1}{c|}{Yes}               & \multicolumn{1}{c|}{Yes}                 & \multicolumn{1}{c|}{Yes}           & Yes               \\ \hline
xyz                & -0.150              & \multicolumn{1}{c|}{7.568094}      & \multicolumn{1}{c|}{6.465555}           & \multicolumn{1}{c|}{\textbf{6.239807}}   & \multicolumn{1}{c|}{6.259547}      & 15.359850         & \multicolumn{1}{c|}{0.011566}      & \multicolumn{1}{c|}{0.009653}          & \multicolumn{1}{c|}{0.009864}            & \multicolumn{1}{c|}{\textbf{0.008517}} & 0.117175          & \multicolumn{1}{c|}{0.126255}          & \multicolumn{1}{c|}{0.126986}          & \multicolumn{1}{c|}{0.148956}            & \multicolumn{1}{c|}{\textbf{0.117847}} & 0.372258          & \multicolumn{1}{c|}{Yes}           & \multicolumn{1}{c|}{Yes}               & \multicolumn{1}{c|}{Yes}                 & \multicolumn{1}{c|}{Yes}           & Yes               \\ \hline
xyz                & 0.000               & \multicolumn{1}{c|}{7.425029}      & \multicolumn{1}{c|}{4.861160}           & \multicolumn{1}{c|}{\textbf{4.859477}}   & \multicolumn{1}{c|}{6.655158}      & 5.883497          & \multicolumn{1}{c|}{0.010969}      & \multicolumn{1}{c|}{\textbf{0.003722}} & \multicolumn{1}{c|}{0.004404}            & \multicolumn{1}{c|}{0.010254}          & 0.026074          & \multicolumn{1}{c|}{0.152020}          & \multicolumn{1}{c|}{\textbf{0.098812}} & \multicolumn{1}{c|}{0.157771}            & \multicolumn{1}{c|}{0.129522}          & 0.379013          & \multicolumn{1}{c|}{Yes}           & \multicolumn{1}{c|}{Yes}               & \multicolumn{1}{c|}{Yes}                 & \multicolumn{1}{c|}{Yes}           & Yes               \\ \hline
xyz                & 0.075               & \multicolumn{1}{c|}{7.115747}      & \multicolumn{1}{c|}{\textbf{4.965403}}  & \multicolumn{1}{c|}{5.040673}            & \multicolumn{1}{c|}{6.277422}      & 7.427389          & \multicolumn{1}{c|}{0.009977}      & \multicolumn{1}{c|}{\textbf{0.003966}} & \multicolumn{1}{c|}{0.004081}            & \multicolumn{1}{c|}{0.008231}          & 0.049423          & \multicolumn{1}{c|}{0.167109}          & \multicolumn{1}{c|}{0.138969}          & \multicolumn{1}{c|}{0.141843}            & \multicolumn{1}{c|}{0.115283}          & 0.376754          & \multicolumn{1}{c|}{Yes}           & \multicolumn{1}{c|}{Yes}               & \multicolumn{1}{c|}{Yes}                 & \multicolumn{1}{c|}{Yes}           & Yes               \\ \hline
xyz                & 0.150               & \multicolumn{1}{c|}{1.359120}      & \multicolumn{1}{c|}{\textbf{4.844337}}  & \multicolumn{1}{c|}{4.875827}            & \multicolumn{1}{c|}{5.126340}      & 0.883666          & \multicolumn{1}{c|}{0.014639}      & \multicolumn{1}{c|}{0.003606}          & \multicolumn{1}{c|}{0.002904}            & \multicolumn{1}{c|}{0.003983}          & 0.001789          & \multicolumn{1}{c|}{0.271770}          & \multicolumn{1}{c|}{0.149187}          & \multicolumn{1}{c|}{0.187735}            & \multicolumn{1}{c|}{0.133928}          & 0.176308          & \multicolumn{1}{c|}{\textbf{No}}   & \multicolumn{1}{c|}{Yes}               & \multicolumn{1}{c|}{\textbf{No}}         & \multicolumn{1}{c|}{Yes}           & \textbf{No}       \\ \hline
z                  & -0.050              & \multicolumn{1}{c|}{7.723198}      & \multicolumn{1}{c|}{\textbf{5.668666}}  & \multicolumn{1}{c|}{6.375411}            & \multicolumn{1}{c|}{8.300653}      & 7.623795          & \multicolumn{1}{c|}{0.019203}      & \multicolumn{1}{c|}{\textbf{0.008097}} & \multicolumn{1}{c|}{0.012004}            & \multicolumn{1}{c|}{0.018782}          & 0.048181          & \multicolumn{1}{c|}{0.121373}          & \multicolumn{1}{c|}{\textbf{0.096880}} & \multicolumn{1}{c|}{0.157460}            & \multicolumn{1}{c|}{0.098883}          & 0.393317          & \multicolumn{1}{c|}{Yes}           & \multicolumn{1}{c|}{Yes}               & \multicolumn{1}{c|}{Yes}                 & \multicolumn{1}{c|}{Yes}           & Yes               \\ \hline
z                  & -0.100              & \multicolumn{1}{c|}{0.015911}      & \multicolumn{1}{c|}{\textbf{10.773465}} & \multicolumn{1}{c|}{11.093853}           & \multicolumn{1}{c|}{11.260336}     & 12.575253         & \multicolumn{1}{c|}{0.000002}      & \multicolumn{1}{c|}{\textbf{0.040536}} & \multicolumn{1}{c|}{0.043189}            & \multicolumn{1}{c|}{0.043243}          & 0.066770          & \multicolumn{1}{c|}{\textbf{0.002792}} & \multicolumn{1}{c|}{0.073959}          & \multicolumn{1}{c|}{0.118678}            & \multicolumn{1}{c|}{0.079583}          & 0.370222          & \multicolumn{1}{c|}{\textbf{No}}   & \multicolumn{1}{c|}{Yes}               & \multicolumn{1}{c|}{\textbf{No}}         & \multicolumn{1}{c|}{Yes}           & Yes               \\ \hline
z                  & 0.000               & \multicolumn{1}{c|}{7.511052}      & \multicolumn{1}{c|}{\textbf{4.978000}}  & \multicolumn{1}{c|}{5.233706}            & \multicolumn{1}{c|}{6.299238}      & 6.076239          & \multicolumn{1}{c|}{0.011097}      & \multicolumn{1}{c|}{\textbf{0.004251}} & \multicolumn{1}{c|}{0.005609}            & \multicolumn{1}{c|}{0.007886}          & 0.027799          & \multicolumn{1}{c|}{0.148883}          & \multicolumn{1}{c|}{0.121492}          & \multicolumn{1}{c|}{0.136734}            & \multicolumn{1}{c|}{\textbf{0.105076}} & 0.402755          & \multicolumn{1}{c|}{Yes}           & \multicolumn{1}{c|}{Yes}               & \multicolumn{1}{c|}{Yes}                 & \multicolumn{1}{c|}{Yes}           & Yes               \\ \hline
z                  & 0.050               & \multicolumn{1}{c|}{6.084520}      & \multicolumn{1}{c|}{\textbf{4.561033}}  & \multicolumn{1}{c|}{4.835022}            & \multicolumn{1}{c|}{5.856955}      & 5.789299          & \multicolumn{1}{c|}{0.006865}      & \multicolumn{1}{c|}{0.003129}          & \multicolumn{1}{c|}{\textbf{0.002743}}   & \multicolumn{1}{c|}{0.006014}          & 0.030311          & \multicolumn{1}{c|}{0.164402}          & \multicolumn{1}{c|}{0.154412}          & \multicolumn{1}{c|}{0.178154}            & \multicolumn{1}{c|}{\textbf{0.143068}} & 0.391134          & \multicolumn{1}{c|}{Yes}           & \multicolumn{1}{c|}{Yes}               & \multicolumn{1}{c|}{Yes}                 & \multicolumn{1}{c|}{Yes}           & Yes               \\ \hline
z                  & 0.100               & \multicolumn{1}{c|}{1.208510}      & \multicolumn{1}{c|}{11.522041}          & \multicolumn{1}{c|}{5.586891}            & \multicolumn{1}{c|}{4.752776}      & 1.591956          & \multicolumn{1}{c|}{0.007685}      & \multicolumn{1}{c|}{0.635960}          & \multicolumn{1}{c|}{0.004139}            & \multicolumn{1}{c|}{0.002017}          & 0.018765          & \multicolumn{1}{c|}{0.168021}          & \multicolumn{1}{c|}{0.143042}          & \multicolumn{1}{c|}{0.199129}            & \multicolumn{1}{c|}{0.154843}          & 0.287403          & \multicolumn{1}{c|}{\textbf{No}}   & \multicolumn{1}{c|}{\textbf{No}}       & \multicolumn{1}{c|}{\textbf{No}}         & \multicolumn{1}{c|}{\textbf{No}}   & \textbf{No}       \\ \hline
\end{tabular}
\end{adjustbox}
\end{table}
\end{landscape}

\section{Reachability Analysis of Single-objective Reinforcement Learning}

This section focuses on further analysis of the single-objective reinforcement learning solutions described in the previous sections. Reinforcement learning-based control can be thought of as an optimal control strategy, which means that the RL agent predicts optimal action given a specific state, and ultimately over time, the agent will meet the reward constraint. This also means that the classical control verification methods can no longer be applied for stability and robustness. However, Hamilton-Jacobi (HJ) reachability analysis is a prominent method when it comes to formal verification of optimal control strategies. \cite{bansal2017hamiltonjacobi} provide an extensive overview of both conventional and contemporary ways of performing reachability analysis. Typically, reachability analysis involves computing a ``reachability set'', given a goal state, a reachability set is a set of all possible states from which the system can reach the given goal state exactly at time $\tau$. A backward reachability tube is defined as the set of all possible states from which the system can reach the goal state within a duration of time $\tau$. If the goal state is a desirable destination, we would like the control policy to have a large set, if on the other hand, the destination is unsafe, we desire our policy to have a minimal set or preferably a null set that can avert the unsafe states. There are several ways of computing these sets for optimal control policies \cite{chen2017decomposition, Kaynama_2013, 4177130, rubiesroyo2017recursive}, that make computation of reachability sets feasible for non-linear systems.

Computing the backward reachability set (BRS) or backward reachability tube (BRT) via the level-set method involves solving the Hamilton-Jacobi-Bellman partial differential equation. Given an initial state $x$ and a time period $\tau$, the Bellman equation can be written as equation \ref{bellman:eq}, where $C$ is the cost function $C: \mathbb{R}^d \times \mathbb{R}^a \to \mathbb{R}$ and $D$ is the terminal cost function $D: \mathbb{R}^d \to \mathbb{R}$. The HJ partial differential equation can be written by Taylor expanding the Bellman equation as in equation \ref{hj_1:eq}, where $\nabla V = \frac{\partial V}{\partial x}$ denotes the spatial derivative, i.e. the gradient of the cost function with respect to the state variables, and $F$ denotes the system dynamics, i.e. the next state given the current state $x$ and action $u$ (derived from the policy $\pi$). If we have an adversarial input or disturbance $d$ acting on the system, we would be interested in minimizing the value function with respect to $u$ while maximizing the cost with respect to the adversarial disturbance $d$ (equation \ref{hj_2:eq}), meaning we would like the control input to minimize the value function making it feasible for the system to reach the goal starting from an initial state, while the adversarial attempts to make it harder for the agent to reach its goal.

\begin{equation}
	\label{bellman:eq}
	V(x, t) = \min_u \int_{t=0}^\tau C(x_t, u_t) dt + D(x_\tau)
\end{equation}

\begin{equation}
	\label{hj_1:eq}
	V(x + dx, t + dt) = \frac{d V(x, t)}{d t} + \min_{u \sim \pi} \nabla V \cdot F(x, u) + C(x, u)
\end{equation}

\begin{equation}
	\label{hj_2:eq}
	V(x + dx, t + dt) = \frac{d V(x, t)}{d t} + \min_{u \sim \pi} \max_{d} \nabla V \cdot F(x, u, d) + C(x, u)
\end{equation}

In our case, the policy $\pi$ is given by a reinforcement learning agent which is already optimized with respect to the reward function (equations \ref{rew1:eq} and \ref{rew2:eq}), so we are only interested in studying the effect of $d$ by looking at the tolerance of the policy $\pi$. To approximate the BST for our RL policy, we can formulate an experiment where $\pi$ emulates an optimal control policy $\pi^*$ and $d$ can then emulate the notion of action errors from the earlier experiments carried out.

Going back to the quadrotor navigation example, where the objective is to navigate and hover at the destination $x,y,z = (0, 0, 1)$. Consider the partial states of the system relevant to the navigation task $p_t = [x_t, y_t, z_t]$ (ignoring the linear and angular velocities which can be assumed to be constant, equation \ref{obs_space:eq}). As before, the control signal or the action is the relative coordinate for the quadrotor to move to, $a_t = u_t = [\Delta x, \Delta y, \Delta z]$. And let $d_t$ denote the deviation or the noise acting on the system. Since the destination coordinate is $(0, 0, 1)$ we can formulate the system dynamics for the optimal control problem by a set of linear equations (equation \ref{opt_c_dyn:eqn}), where the reference signal $r = [r_x, r_y, r_z] = [0, 0, 1]$ and $d_x, d_y, d_z$ are assumed to be Gaussian noise. Equation \ref{opt_c_dyn:eqn} represents the $F(x, u, d)$ term in the Hamilton-Jacobi-Bellman partial differential equation (equation \ref{hj_2:eq}). The term $\nabla V \cdot F$ can further be decomposed as $\nabla V \cdot F (x, u, d)= \nabla V \cdot F_1(x, u) + \nabla V \cdot F_2(x, d)$ as the disturbances act additively on the system (equation \ref{opt_c_dyn:eqn}). This yields equation \ref{hj_3:eq}, where $F_1$ is the ideal dynamics of the system given the state and action, while $F_2$ is the dynamics capturing the disturbance.

\begin{align}
	\label{opt_c_dyn:eqn}
	x_{t+1} = r_x - x_t + \Delta x + d_x  = 0 - x_t + \Delta x + d_x \notag \\
	y_{t+1} = r_y - y_t + \Delta y + d_y = 0 - y_t + \Delta y + d_y \notag \\
	z_{t+1} = r_z - z_t + \Delta z + d_z = 1 - z_t + \Delta z + d_z
\end{align}

\begin{equation}
	\label{hj_3:eq}
	V(x + dx, t + dt) = \frac{d V(x, t)}{d t} + \min_{u \sim \pi} \nabla V \cdot F_1(x, u) + \max_{d} \nabla V \cdot F_2(x, d) + C(x, u)
\end{equation}

The RL policy $\pi$ was trained to maximize the negative of the distance from the reference point (equation \ref{rew1:eq}). Thus, the value function $V$ can simply represent the distance from the reference point $r = [r_x, r_y, r_z]$. The solution to the optimal control problem (equation \ref{opt_c_dyn:eqn}) is $(x_t, y_t, z_t) = (r_x, r_y, r_z)$. As the value function is known, we can use an optimal bang-bang control policy $\pi^*$ to emulate the optimal RL policy $\pi$ and evaluate the robustness by imposing the deviations $d$ experienced by the different RL agents. This would make it feasible to approximate the backward reachability tube using the classic level-set method. The assumption we make here is that the optimal RL policy under no disturbance is similar to a bang-bang control policy, which is valid because the bang-bang policy optimizing the HJ PDE follows the optimal path and so does the RL agent when no disturbances are acting on the system.

\begin{figure*}[h]
	\centering
	\caption{Bang-bang Policy}
	\begin{subfigure}[b]{0.35\textwidth}
		\caption{Optimal Controls XY plane. Destination $(0, 0)$}
		\includegraphics[width=\textwidth]{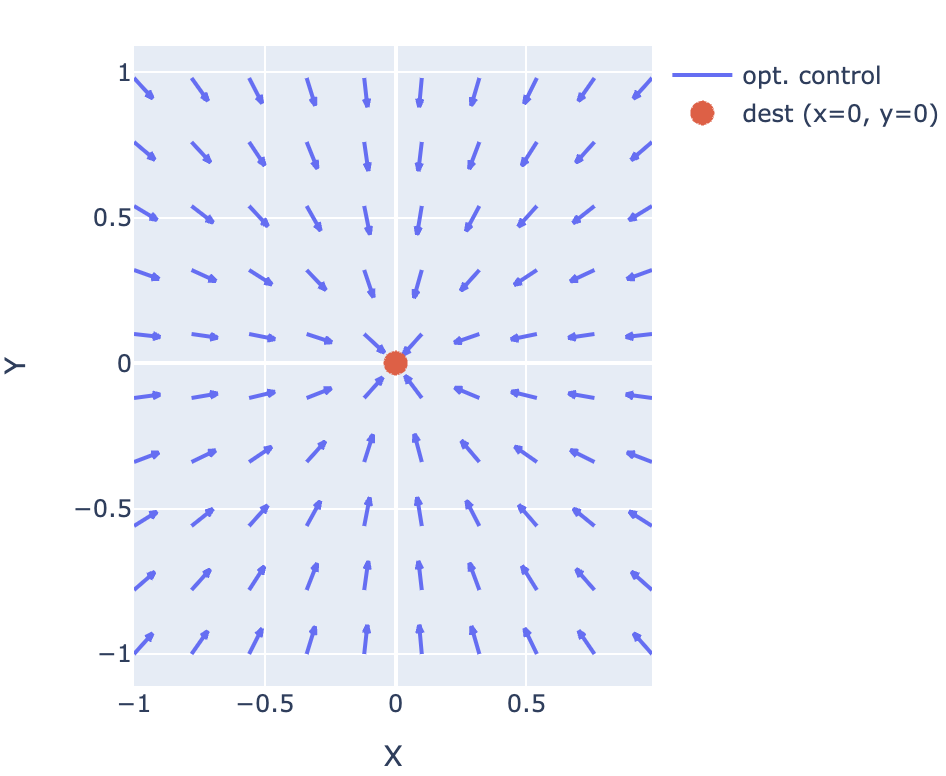}
		\label{opt_u_xy:fig}
	\end{subfigure}
	\hfill
	\begin{subfigure}[b]{0.35\textwidth}
		\caption{Optimal Controls XZ plane. Destination $(0, 1)$}
		\includegraphics[width=\textwidth]{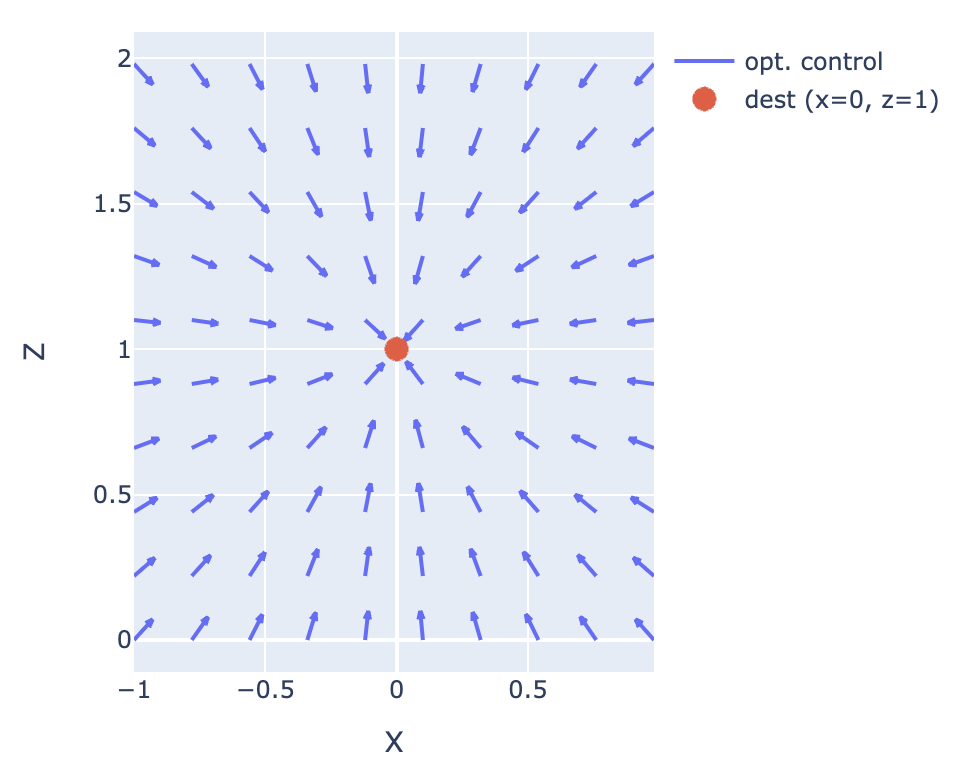}
		\label{opt_u_xz:fig}
	\end{subfigure}
\end{figure*}

Given the state $p_t = [x_t, y_t, z_t]$, the value function is given by equation \ref{vf:eq}. The spatial gradients of the value function is the partial derivatives with respect to the state variables (equation \ref{vfp:eq}). A bang-bang control policy to minimize the HJ PDE will be to use a step in the opposite direction of the spatial gradient (equation \ref{opt_u:eq}), and the optimal disturbance inhibiting the agent from reaching its destination or attempting to maximize the HJ PDE is given by equation \ref{opt_d:eq}. Figures \ref{opt_u_xy:fig} and \ref{opt_u_xz:fig} visualize the optimal actions, we observe that the bang-bang policy drives the agent to the destination $(0, 0, 1)$.

\begin{equation}
	\label{vf:eq}
	V(p_t) = V([x_t, y_t, z_t]) = \sqrt{x_t^2 + y_t^2 + (1 - z_t)^2}
\end{equation}

\begin{equation}
	\label{vfp:eq}
	\nabla V(p_t) = 
	\begin{bmatrix}
	\frac{\partial V}{\partial x} \\
	\frac{\partial V}{\partial y} \\
	\frac{\partial V}{\partial z} \\
\end{bmatrix} = \frac{1}{\sqrt{x_t^2 + y_t^2 + (1 - z_t)^2}}
\begin{bmatrix}
	x_t \\
	y_t \\
	-(1 - z_t)
\end{bmatrix}
\end{equation}

\begin{equation}
	\label{opt_u:eq}
	u_i = \begin{cases}
		-|\Delta u_{max}| \ \text{if} \frac{\partial V}{\partial x_i} > 0 \\
		|\Delta u_{max}| \ \text{if} \frac{\partial V}{\partial x_i} < 0
	\end{cases}
\end{equation}

\begin{equation}
	\label{opt_d:eq}
	d_i = \begin{cases}
		|\Delta d_{max}| \ \text{if} \frac{\partial V}{\partial x_i} > 0 \\
		-|\Delta d_{max}| \ \text{if} \frac{\partial V}{\partial x_i} < 0
	\end{cases}
\end{equation}

The next objective is then to model the disturbances $d$ for the linearized system. This disturbance should emulate what the RL agent experiences during the evaluation phase. For the particular experiment, we pick a step wind disturbance (of magnitude 0.075 N along $XYZ$) acting on the quadrotor during the episode and simply capture the errors during the flight, the error is just the difference between the intended position of the quadrotor by executing the action and where it really ended up, i.e. $x_{t+1} - (x_t + u_t)$. The statistics of these errors are listed in tables \ref{err1_base:table} and \ref{err2_dist_err:table}, clearly the covariances of the dist-err model (proposed model) are significantly lower, this essentially means that the model was able to execute its actions more concretely compared to the baseline model. We will now use these disturbances (1 standard deviations apart) as $d_x, d_y, d_z$ while computing the BRT. This would reveal the worst-case tolerances of the models and enable us to compare the robustness.

\begin{table}[h]
\centering
\caption{Error Statistics - baseline model}
\label{err1_base:table}
\begin{tabular}{|c|c|c|c|c|}
\hline
\textbf{} & \multicolumn{3}{c|}{\textbf{Covariances}} & \textbf{Means} \\ \hline
          & $x_e$        & $y_e$        & $z_e$       &                \\ \hline
$x_e$     & 0.05963      & 0.02477      & 0     & $-0.01970$    \\ \hline
$y_e$     & 0.02477      & 0.09416      & 0.01642     & $-0.0271$    \\ \hline
$z_e$     & 0      & 0.01642      & 0.08471     & $0.01089$    \\ \hline
\end{tabular}
\end{table}

\begin{table}[h]
\centering
\caption{Error Statistics - \textbf{dist-err} model}
\label{err2_dist_err:table}
\begin{tabular}{|c|c|c|c|c|}
\hline
\textbf{} & \multicolumn{3}{c|}{\textbf{Covariances}} & \textbf{Means} \\ \hline
          & $x_e$        & $y_e$        & $z_e$       &                \\ \hline
$x_e$     & 0.03474      & 0      & 0     & $-0.02000$    \\ \hline
$y_e$     & 0      & 0.02833      & 0.00691     & $0.03506$    \\ \hline
$z_e$     & 0      & 0.00691      & 0.02764     & $0.010255$    \\ \hline
\end{tabular}
\end{table}

\begin{figure*}[h]
	\centering
	\caption{Backward Reachability Tubes (BRT). \newline Models compared: i. \textbf{baseline}: the baseline model. ii. \textbf{dist-err}: the proposed arch for single-objective, receives feedback for the action errors.}
	\begin{subfigure}[b]{0.3\textwidth}
		\caption{3D BRT baseline model}
		\includegraphics[width=\textwidth]{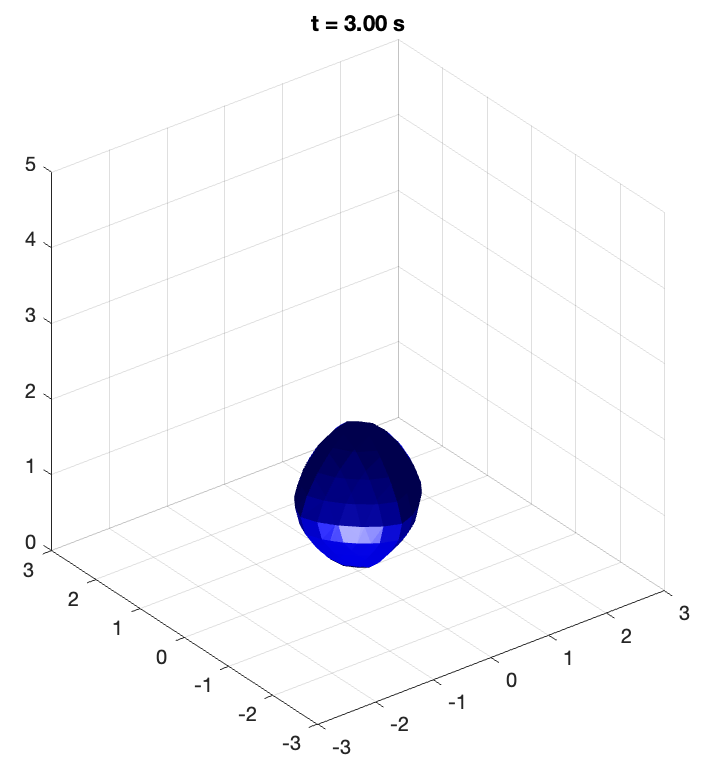}
		\label{brt_base:fig}
	\end{subfigure}
	\hfill
	\begin{subfigure}[b]{0.3\textwidth}
		\caption{3D BRT \textbf{dist-err} model}
		\includegraphics[width=\textwidth]{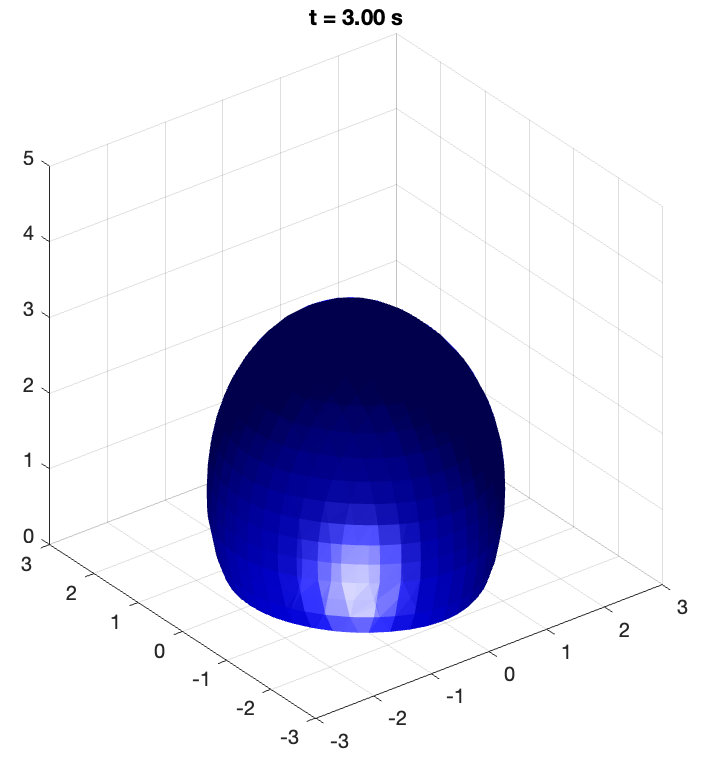}
		\label{brt_dist_err:fig}
	\end{subfigure}
	\hfill
	\begin{subfigure}[b]{0.3\textwidth}
		\caption{Comparison of BRTs, 2D slice ($z = 1$)}
		\includegraphics[width=\textwidth]{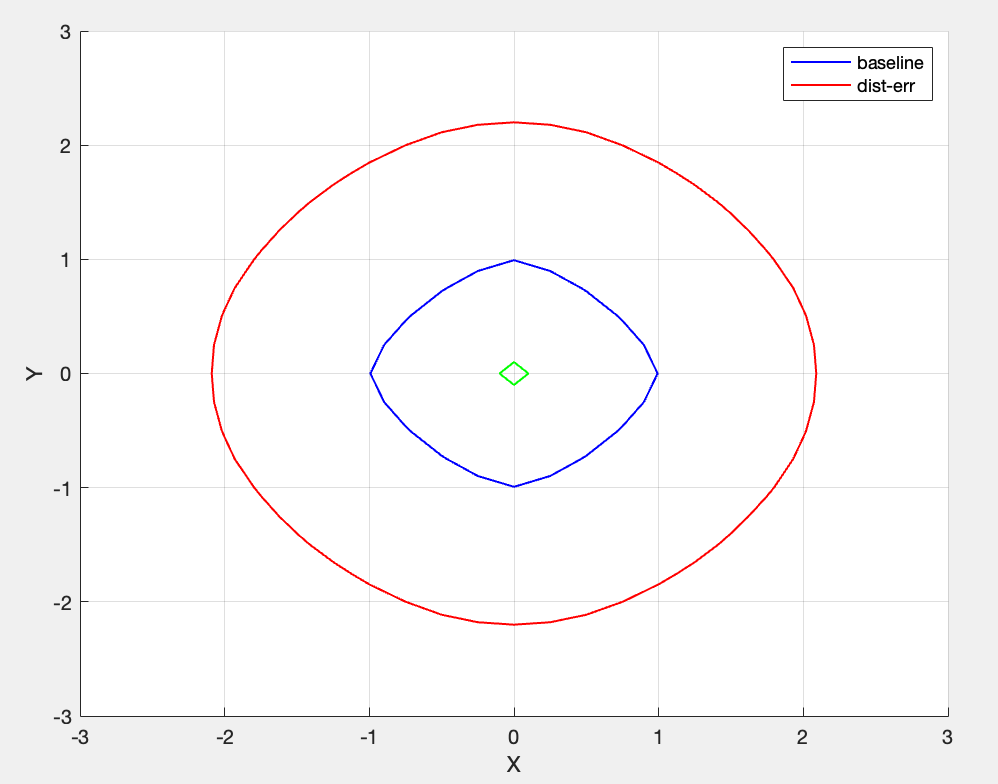}
		\label{brts_cmp:fig}
	\end{subfigure}
\end{figure*}

The target set is a sphere of radius 10 cm centered at $[0, 0, 1]$. As we are interested in evaluating the ability of the agent to reach the target set, the BRT computed at the end will comprise of all the possible initial states from which the agent can navigate to the specified target set. The time duration for computing the BRT is set to $\tau = 3 \ \text{s}$, meaning we look backward starting from our target set for all the possible reachable sets within a duration of 3s. The robustness of the proposed model is clearly evident (figures \ref{brt_base:fig} and \ref{brt_dist_err:fig}). The reachability tube of the proposed model is much larger, implying that the agent can reach the target set from a wider range of initial states within the span of 3 seconds. Figure \ref{brts_cmp:fig} compares the 2D slices of the BRTs at $z = 1$. Although these tubes are a crude approximation due to the fact that the optimal RL policy was emulated by an optimal bang-bang policy, we get a tangible measure of robustness by computing the BRTs. These results further corroborate the observations from the previous section. For instance, the ``converged'' column in table \ref{metrics:table} represents the true simulations of the trained RL agents, and we see that the baseline model fails to converge in two additional cases where the proposed model (dist-err) has no problem reaching the target.

%
%

\section{Multi-objective Reinforcement Learning}

Inducing the notion of conservativeness can have a positive impact on interpretability, but it may not be always desired for the model to act conservatively. In certain applications, the agent may be expected to take a risk and achieve the best RL reward, on the other hand, in the context of safety-critical applications, we would want the agent to adhere to certain physical constraints. In this regard, it would be desirable to have the ability to tune the level of conservativeness required and dynamically change it during the runtime. So a multi-objective RL problem can be formulated where the primary objective is task-specific and additional reward objectives can be introduced to constrain the agent. We would then employ a scalarized expected reward (SER) approach to learn an array of RL policies each of which prioritizes the rewards differently. A typical SER reward for multi-objective RL similar to equation \ref{multi_rl:eq}, where $w \in W$ can be a finite set of weight vectors for the agents to weigh the rewards. Thus, we can learn an array of policies that can differently prioritize the conservativeness $R_{err}$. And the key difference to the single-objective RL task (equation \ref{utility_fun:eq}) is that the utility function would be applied after the expectation.

\begin{equation}
	\label{multi_rl:eq}
	U \left(
	\mathbb{E}\begin{bmatrix}
	R_{nav} \\
	R_{err}
	\end{bmatrix}
	\right) = w^T \begin{bmatrix}
	\mathbb{E} \left[ R_{nav} \right] \\
	\mathbb{E} \left[R_{err} \right]
	\end{bmatrix}
\end{equation}

\section{Connection to Model-based Reinforcement Learning}

All the results discussed thus far required formulating a reward objective that would penalize the agent for deviating from the normal or expected behavior. The combination of hierarchical control and the specific way of formulating the actions serendipitously made it feasible to compose an error model without any additional effort. However, in cases where such a formulation is not as obvious, we can leverage model-based reinforcement learning methods. Typically, the model-based reinforcement learning methods learn the dynamics of the system by fitting a non-linear model (often a neural network) that can predict the next state given the current state, i.e. $s_{t+1} = \hat{f}(s_t, a_t)$ and leverage it later on in several ways. By building up a dataset $\mathcal{D}$ comprising of observations $(s_t, a_t, s_{t+1})$ under ideal conditions, we can thus learn the ``ideal'' dynamics of the system. Once $\hat{f}$ is learned, the rest of the procedure would be exactly the same, i.e. introducing a reward objective for the system to behave conservatively as per the ideal dynamics. When the dynamics span a high-dimension space, it should be discretionary to choose a subset of states that make sense, as in the task of quadrotor discussed in the paper, although the actual state-space of the quadrotor spanned 15 dimensions, only 3 of the states were absolutely vital for interpretability. Thus, we could formulate a reward objective for such states.

\section{Conclusion}

Robustness and interpretability are some of the key challenges for contemporary reinforcement learning-based applications, and the paper presents a plausible approach to alleviate the issues in the context of standard reinforcement learning. The approach of introducing a secondary reward objective constraining the dynamics of the agent was found to have a positive impact after rigorous testing and comparison. Although the constrained agent the new objective did not necessarily outperform in terms of the RL reward, it was observed that the agent exhibited some very nice physical properties under evaluation and was found to have an edge when looking at a relevant set of physical metrics for the task involved.

It is important to note that the approach discussed does guarantee that the process dynamics are adhered to always, unlike MPC-based solutions that guarantee a feasible solution, this is a distinguishing quality of model predictive control. However, MPC has its own set of pros and cons compared to reinforcement learning. This work attempts to improve and elevate the problems surrounding classical RL in optimal control-based applications. While the task considered was restricted to a navigation problem of a quadrotor, the ideas presented can be extended to any non-linear control problem, formulating a hierarchical control task with a partial dynamics model would be ideal, in the case where the dynamics are known, a non-linear model could be fit on the data to estimate the crucial states for the problem, which could then serve as a ground truth for the RL task.

\bibliographystyle{apalike}
\bibliography{./resources/bib.bib}

\end{document}